\documentclass[journal]{IEEEtran}

\usepackage{amsthm}
\usepackage{amsfonts}
\usepackage{subfig}
\usepackage{multirow}
\usepackage{graphicx}
\usepackage{lscape}
\usepackage{rotating} 
\usepackage{arydshln}	
\usepackage{wrapfig} 

\usepackage{hyperref}
\usepackage{url}
\usepackage{setspace}

\usepackage{makecell}
\usepackage{stfloats}

\usepackage{booktabs}
\usepackage[table,xcdraw,dvipsnames]{xcolor}

\definecolor{burntorange}{HTML}{CC5500}

\usepackage{cite}
\usepackage[numbers,sort&compress]{natbib}

\usepackage{algorithm}
\usepackage{algpseudocode}
\usepackage{amsmath}

\begin{document}

\title{Ask-AC: An Initiative Advisor-in-the-Loop Actor-Critic Framework}

\author{Shunyu~Liu,
        Kaixuan~Chen,
        Na~Yu,
        Jie~Song,
        Zunlei~Feng,
        Mingli~Song
\thanks{This article has been accepted for publication by IEEE Transactions on Systems, Man and Cybernetics: Systems. The published version is available at \url{https://doi.org/10.1109/TSMC.2023.3296773}. This work was supported in part by the National Key Research and Development Program of China under Grant 2018AAA0101503, and in part by the Science and Technology Project of State Grid Corporation of China (SGCC): Fundamental Theory of Human-in-the-Loop Hybrid- Augmented Intelligence for Power Grid Dispatch and Control. \textit{(Corresponding author: Mingli Song.)}}
\thanks{Shunyu Liu, Kaixuan Chen, Na Yu, and Mingli Song are with the College of Computer Science and Technology, Zhejiang University, Hangzhou 310027, China (e-mail: liushunyu@zju.edu.cn; na\_yu@zju.edu.cn; chenkx@zju.edu.cn; brooksong@zju.edu.cn).}
\thanks{Jie Song and Zunlei Feng are with the College of Software Technology, Zhejiang University, Hangzhou 310027, China (e-mail: sjie@zju.edu.cn; \text{zunleifeng}@zju.edu.cn).}
\thanks{Digital Object Identifier 10.1109/TSMC.2023.3296773}
}

\markboth{IEEE TRANSACTIONS ON SYSTEMS, MAN, AND CYBERNETICS: SYSTEMS}
{Liu \MakeLowercase{\textit{et al.}}: Ask-AC: An Initiative Advisor-in-the-Loop Actor-Critic Framework}

\IEEEpubid{\begin{minipage}{\textwidth}\ \\[30pt] \centering 2168-2216~\copyright~2023 IEEE. Personal use is permitted, but republication/redistribution requires IEEE permission.\\See https://www.ieee.org/publications/rights/index.html for more information.\end{minipage}}

\maketitle

\begin{abstract}
  Despite the promising results achieved, state-of-the-art interactive reinforcement learning schemes rely on passively receiving supervision signals from advisor experts, in the form of either continuous monitoring or pre-defined rules, which inevitably result in a cumbersome and expensive learning process. In this paper, we introduce a novel \emph{initiative} advisor-in-the-loop actor-critic framework, termed as Ask-AC, that replaces the unilateral advisor-guidance mechanism with a bidirectional learner-initiative one, and thereby enables a customized and efficacious message exchange between learner and advisor. At the heart of Ask-AC are two complementary components, namely action requester and adaptive state selector, that can be readily incorporated into various discrete actor-critic architectures. The former component allows the agent to initiatively seek advisor intervention in the presence of uncertain states, while the latter identifies the unstable states potentially missed by the former especially when environment changes, and then learns to promote the ask action on such states. Experimental results on both stationary and non-stationary environments and across different actor-critic backbones demonstrate that the proposed framework significantly improves the learning efficiency of the agent, and achieves the performances on par with those obtained by continuous advisor monitoring. 
\end{abstract}


\IEEEpeerreviewmaketitle

\section{Introduction}

Deep reinforcement learning focuses on extracting features from input states and delivering response actions in an end-to-end fashion~\cite{books:books/lib/SuttonB98}. In recent years, this learning paradigm has witnessed its unprecedented success in many game-based tasks~\cite{DQN15,AlphaGo,AlphaStar} and robot-based tasks~\cite{SAC,DDPG}. Despite the encouraging results accomplished, methods along this line are often prone to low sampling efficiency, since the agent requires a large amount of training data to explore the highly-complex environment.

\begin{figure}[!t]
  \centering
    \includegraphics[scale=1.05]{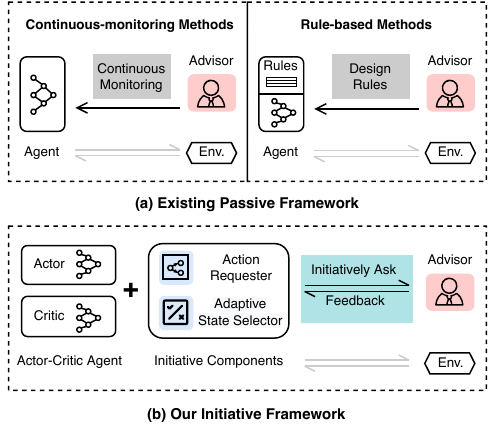}
    \vspace{-0.15cm}
    \caption{Comparing the proposed initiative Ask-AC framework with the existing passive framework for interactive reinforcement learning. Unlike the passive framework with unidirectional interactions, Ask-AC introduces a bidirectional message exchange scheme to endow the agent with an initiatively-asking mechanism, which largely reduces the demand for advisor feedback. ``Env.'' denotes environment.}
    \vspace{-0.8cm}
    \label{fig:related}
\end{figure}

\IEEEpubidadjcol

To alleviate the sampling efficiency issue, interactive reinforcement learning methods~\cite{DBLP:conf/ijcai/Taylor18,DBLP:conf/ijcai/ZhangTGBS19,DBLP:conf/ijcai/CederborgGIT15,DBLP:conf/nips/GriffithSSIT13,DBLP:conf/kcap/KnoxS09,DBLP:conf/icml/MacGlashanHLPWR17,DBLP:conf/aaai/WarnellWLS18,yang2022optimal,DBLP:conf/ijcai/GoyalNM19,ProLoNets,DBLP:journals/nca/Treesatayapun20,DBLP:conf/ijcai/ZhangHWTMDZ20,zhou2022smart} have been proposed as a competent alternative. State-of-the-art interactive approaches rely on unidirectional guidance that passively collects supervision signals from advisor experts~(\emph{i.e.} humans or trained agents). Specifically, these approaches can be broadly categorized into two classes: continuous-monitoring methods and rule-based ones, as depicted in the upper row of Figure~\ref{fig:related}. Continuous-monitoring methods~\cite{DBLP:conf/ijcai/CederborgGIT15,DBLP:conf/nips/GriffithSSIT13,DBLP:conf/kcap/KnoxS09,DBLP:conf/icml/MacGlashanHLPWR17,DBLP:conf/aaai/WarnellWLS18,yang2022optimal} demand advisor experts to constantly monitor the learner and provide supervisions, yielding an excessively heavy training process. Furthermore, the amount of supervision is often limited by human availability or agent communication cost, hence the agent must consider the burden of advisor experts to reduce interaction. Rule-based methods~\cite{DBLP:conf/ijcai/GoyalNM19,ProLoNets,DBLP:journals/nca/Treesatayapun20,DBLP:conf/ijcai/ZhangHWTMDZ20,zhou2022smart}, on the other hand, impose action rules before training to guide the agent, so that advisers are no longer required to participate during training. Nevertheless, defining rules prior to training is inevitably very challenging in the case of complex environments; moreover, the hand-crafted rules often lack adaptivity to the task of interest and hence give rise to inferior performances.

In this paper, we introduce a novel initiative interactive reinforcement learning framework, termed as Ask-AC, to reduce the demand for advisor attention while maintaining a high sampling efficiency. Unlike prior interactive methods that merely acquire advisor responses in a passive manner, the proposed Ask-AC, as depicted in the lower row of Figure 1, replaces the unilateral advisor-guidance mechanism with a bidirectional learner-initiative one. In other words, Ask-AC enables the agent to \emph{initiatively} seek supervision signals from advisor experts only when the agent confronts uncertain states, and hence significantly alleviates advisor efforts as compared to the continuous-monitoring scheme. Besides, the two-way message exchange allows for adaptive and timely advisor supervision, especially in complex environments, where the rule-based methods are prone to failure since working rules are arduous to design.

The proposed Ask-AC comprises two key components, namely action requester and adaptive state selector, which complement each other and together can be seamlessly integrated with various discrete actor-critic architectures. The action requester endows the agent with a novel initially-asking action, allowing the agent to seek feedback from advisor experts in the presence of uncertain states. Nevertheless, the demand for seeking advisor feedback, as the training progresses, is gradually inhibited by the loss of the action requester. As a result, the agent may fail to perceive the critical states that require advisor feedback especially when the environment changes. To this end, the adaptive state selector is introduced to promote the asking action and to rapidly adapt to the environmental change. This is achieved through analyzing the error of state values and identifying the unstable states potentially missed by the action requester in its history states, so as to acquire advisor guidance.

Our contribution is therefore a dedicated \emph{initiative} advisor-in-the-loop actor-critic framework, which enables a two-way message passing and seeks advisor assistance only on demand. The proposed Ask-AC substantially lessens the advisor participation effort and is readily applicable to various discrete actor-critic architectures. Experimental results on both stationary and non-stationary environments demonstrate that Ask-AC significantly improves the learning efficiency of the agent, and in practice, leads to performances on par with those achieved by continuous advisor monitoring. Moreover, the robustness analysis experiment also shows the effectiveness of our method in the case of advisers with various operation levels.

\section{Related Work}

We briefly review here some topics that are most related to the proposed work, including interactive reinforcement learning, policy distillation and learning from demonstration.

\subsection{Interactive Reinforcement Learning}

Interactive reinforcement learning aims to use additional supervision signals from advisor experts to alleviate the sampling efficiency issue in reinforcement learning, which has attracted increasing attention. Existing interactive reinforcement learning methods can be mainly categorized into two classes: continuous-monitoring methods and rule-based ones. These methods rely on the passive advisor-guidance framework, and thereby inevitably resulting in a cumbersome and expensive learning process.

From the perspective of the continuous-monitoring methods, references~\cite{DBLP:conf/ijcai/CederborgGIT15,DBLP:conf/nips/GriffithSSIT13} proposed a Bayesian approach for integrating the binary advisor feedback on the correctness of actions to shape agent policy. Moreover, \citet{DBLP:conf/icml/MacGlashanHLPWR17} focused on accelerating learning by constant policy-dependent advisor feedback, indicating how much an action improved the performance of the current policy. References~\cite{DBLP:conf/kcap/KnoxS09,DBLP:conf/aaai/WarnellWLS18} leveraged the scalar-valued feedback from real-time advisor interaction to facilitate the learning of the agent. Despite the promising results on improving sampling efficiency achieved, these methods require advisers to monitor the behavior of the agent continuously, which causes a heavy time burden on human advisers or an expensive communication cost on agent advisers. In contrast to the continuous-monitoring methods, our framework introduces a bidirectional learner-initiative mechanism. This mechanism enables the agent to initiatively seeks advisor assistance only on demand, thus significantly relieves advisers from exhaustively being involved in the learning process.

On the other hand, rule-based methods use pre-defined rules to replace advisers to guide the agent, so that advisers are no longer required to participate during training. \citet{DBLP:conf/ijcai/GoyalNM19} used natural language rules to generate auxiliary reward, which was conducive to the guide the exploration behavior of agent under the sparse reward setting. Similarly, \citet{ProLoNets} encoded the advisor domain knowledge rules into a neural decision tree to warm start the learning process. Furthermore, references~\cite{DBLP:journals/nca/Treesatayapun20,DBLP:conf/ijcai/ZhangHWTMDZ20} proposed a knowledge guided policy network framework that uses fuzzy rules to combine prior advisor knowledge with reinforcement learning. However, the rule-based methods often suffer from the problem that rules are hard to design, especially in complex environments, while our framework allows for adaptive and timely message exchange between agent and advisor.

To achieve the same purpose, several action advising methods in interactive reinforcement learning also studied how to reduce the interaction demand~\cite{DBLP:journals/jair/SilvaC19}. However, most of these methods assumed that the agent could accesse and utilize the latent model of advisers, which is inconsistent with our setting. References~\cite{DBLP:conf/ijcai/AmirKKG16,DBLP:conf/atal/TorreyT13} assumed that the advisor used Q network and calculated the state importance by the output of advisor network to determine when to give action advice. References~\cite{DBLP:conf/atal/SilvaGC17,DBLP:conf/aaai/OmidshafieiKLTR19} introduced a new paradigm for learning to teach in cooperative multi-agent settings, where agents can simultaneously learn and advise each other. However, in our setting, the latent model of advisers can not be accessed and utilized, especially for human advisers. The advisers can only return action feedback according to the given state, which is more restrictive than these methods.
To this end, \citet{DBLP:conf/ijcai/AmirKKG16} also proposed a heuristic method to compute the difference of the Q value estimates from agent network and ask for help when the difference was below the pre-defined threshold. Similarly, \citet{DBLP:conf/aaai/SilvaHKT20a} sought advisor intervention only when the variance of the Q value estimates was above the pre-defined threshold. All these heuristic methods depend on the setting of prior threshold, which numerical magnitude is directly related to the Q value of different tasks. Therefore, it is tough to define the threshold in advance without massive experimental evaluation, while the initially-asking ability of our framework can be learned by the agent and does not rely on a pre-defined crucial threshold.

\subsection{Policy Distillation \& Learning from Demonstration}
The other two vital research areas that leverage advisor knowledge in reinforcement learning are policy distillation in transfer learning~\cite{DBLP:journals/corr/ParisottoBS15,DBLP:journals/corr/RusuCGDKPMKH15,DBLP:conf/nips/TehBCQKHHP17,DBLP:conf/aaai/YinP17,DBLP:journals/jair/SilvaC19} and learning from demonstration in imitation learning~\cite{DBLP:journals/ras/ArgallCVB09,DBLP:conf/aaai/HesterVPLSPHQSO18,DBLP:conf/nips/HoE16,DBLP:journals/csur/HusseinGEJ17,DBLP:conf/icml/KangJF18,DBLP:conf/nips/KimFPP13,DBLP:conf/ijcai/TorabiWS18,DBLP:conf/ijcai/WangT19}. Policy distillation focuses on transferring the knowledge by using the action distribution of advisor policy, which cannot be accessed and utilized in our setting. Learning from demonstration aims to learn from the pre-prepared demonstration data, whereas collecting high-quality advisor data could be expensive, and the generalization to unseen data is limited. \citet{DBLP:conf/ijcai/WangT19} proposed a confidence-based method to ask for more demonstration when training, but the confidence calculation was also based on the massive data prepared before training. Unlike learning from demonstration, our advisor-in-the-loop reinforcement learning framework focuses on leveraging advisor knowledge by directly interacting with advisor and receiving real-time feedback based on the current state.

\section{Preliminary}

In this paper, we assume that the latent model of advisor can not be accessed and advisor can only return action feedback according to the given state, due to the limitation of human availability or agent communication cost~\cite{DBLP:conf/ijcai/AmirKKG16,DBLP:conf/aaai/SilvaHKT20a,DBLP:conf/atal/TorreyT13}. Following this assumption, we focus on tasks of discrete action space in both the stationary and the non-stationary environments. 

\subsection{Markov Decision Process}
Firstly, we consider the stationary environment model, which is defined as a Markov decision process (MDP). A MDP is represented by a tuple $\mathcal{M} = <\mathcal{S,A},\mathcal{P},\mathcal{R}, \gamma>$, where $\mathcal{S}$ is the set of continuous states, $\mathcal{A}$ is the set of discrete actions, $\mathcal{P}: \mathcal{S} \times \mathcal{A} \to \mathcal{S}$ is the state transition function, $\mathcal{R}: \mathcal{S} \times \mathcal{A} \times \mathcal{S} \to \mathbb{R} $ is the reward function, and $\gamma \in [0,1]$ is the discount rate. At each discrete time step $t$, the agent can choose the action $a \in \mathcal{A}$ based on the policy $\pi: \mathcal{S} \to \mathcal{A}$ at the given state $s \in \mathcal{S}$. Then the agent will observe the next state of the environment $s'$ with a reward signal $r$ according to the transition function $\mathcal{P}(s' | s, a)$ and the reward function $\mathcal{R}(s,a)$. The objective of the reinforcement learning agent is to learn an optimal policy $\pi^*$ that maximize the expected return $\mathbb{E}_{\pi}[G_t] = \mathbb{E}_{\pi}[\sum_{i=0}^{\infty}{\gamma^{i} r_{i + t}}]$.

The non-stationary environment model can be defined as a sequence of MDPs as $\{\mathcal{M}_1,...,\mathcal{M}_K\}$, where $K$ is the length of the MDPs sequence. For each $k \in [1,K]$, $\mathcal{M}_k = <\mathcal{S,A},\mathcal{P}_k,\mathcal{R}_k, \gamma>$. $\{T_1,...,T_{K-1}\}$, an increasing sequence of random integers, is the changepoint set of the non-stationary environment model. At time $T_k$, the environment model will change from $\mathcal{M}_{k}$ to $\mathcal{M}_{k+1}$. The goal of agent is to learn to rapidly adjust its policy $\pi$ and adapt to the new MDP when environment changes.

\subsection{Actor-Critic Architecture}
In policy-based model-free reinforcement learning methods, the parameterized policy $\pi_\theta$ can be learned by performing gradient ascent on the expected return $\mathbb{E}_{\pi_\theta}[G_t]$ to update the parameters $\theta$. One of the classic policy-based methods is REINFORCE~\cite{REINFORCE}, which updates the policy parameters $\theta$ in the direction $G_t \nabla_\theta \ln{\pi_\theta(a_t|s_t)}$. Moreover, to reduce the high variance of gradient estimation while keeping the bias unchanged, a general method is to subtract an estimated state value baseline $V_{\psi }(s_t)$ from return~\cite{books:books/lib/SuttonB98}. The value function $V_{\psi}$ is parameterized by $\psi$. This method can be viewed as an actor-critic~(AC) architecture~\cite{barto2020looking} where the policy $\pi_\theta$ is the actor and the value function $V_{\psi}$ is the critic. Nowadays, actor-critic architecture is one of the most commonly used mainstream methods in reinforcement learning~\cite{SAC,DDPG,A3C,PPO}, and thereby we aim to propose a framework that can be readily incorporated into various discrete actor-critic architectures~\cite{A3C,PPO}.

\section{Method}

\begin{figure*}[!t]
  \centering
  \includegraphics[width=\textwidth]{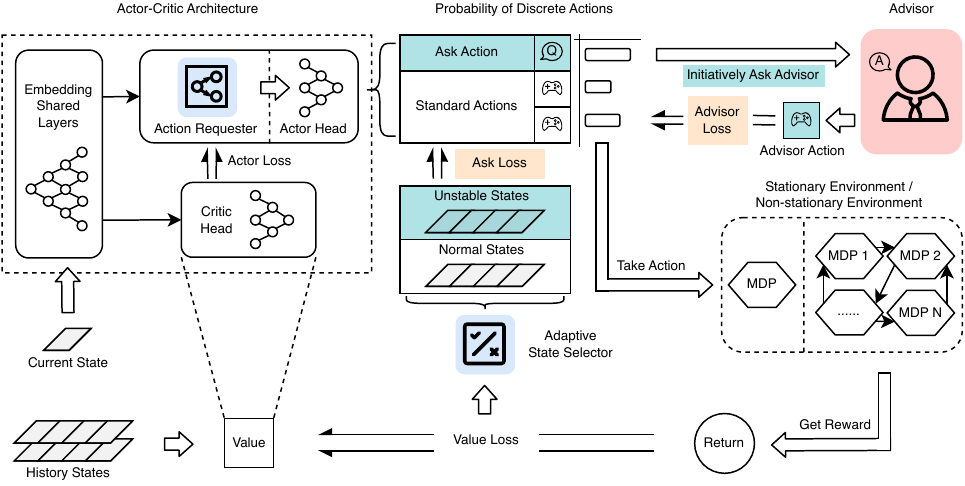}
  \vspace{-0.2cm}
  \caption{Illustration of the proposed Ask-AC framework. Ask-AC is an initiative advisor-in-the-loop actor-critic framework that comprises two complementary components, namely action requester and adaptive state selector.}
  \label{fig:framework}
  \vspace{-0.6cm}
\end{figure*}

In what follows, we detail the proposed initiative advisor-in-the-loop actor-critic framework, termed as Ask-AC, to realize a bidirectional learner-initiative mechanism. As shown in Figure~\ref{fig:framework}, our framework consists of two complementary components: action requester and adaptive state selector. 
The action requester is directly based on a binary classification model with ask action, which enables the agent to initiatively ask for advisor feedback when it encounters uncertain states. 
As a complement to the action requester, the adaptive state selector analyzes the trend of value loss to identify the unstable states that the action requester misses in history states, so as to learn to promote the ask action on such states. With these two components, we can transform the various discrete actor-critic architecture into our advisor-in-the-loop setting.

\subsection{Action Requester}

The conventional interaction problem involves seeking advisor feedbacks at every state which can be computationally expensive. We now design an action requester which allows the agent to learn to ask only in some uncertain states in which advisor feedbacks are most useful. To achieve this, the interaction problem is formalized as a binary classification problem. Considering the discrete action set $\mathcal{A}$ of the actor $\pi_{\theta}$, we introduce an additional action set $\mathcal{A}^{+} = \{A_{ask}, A_{exec}\}$ including the asking action $A_{ask}$ and execution action $A_{exec}$. The implementation of the action requester is a binary classifier $g_{\phi}$ parameterized by $\phi$ based on the additional action set $\mathcal{A}^{+}$. At each discrete time $t$, the agent first makes a binary decision to decide to whether to ask advisor or not based on the action requester $g_{\phi}$. If the action requester chooses the asking action $A_{ask}$ when receiving a state $s \in \mathcal{S}$, the agent will send the state to the advisor. The state $s$ requiring intervention can be considered as an uncertain state. Then the advisor will provide the advisor action feedback $a^{*} \in \mathcal{A}$ according to their prior experience and knowledge. After that, the agent will take the advisor action $a^{*}$ in the environment. On the contrary, the agent will directly execute an action $a \in \mathcal{A}$ based on its own actor network $\pi_{\theta}$ if the action requester chooses the execution action $A_{exec}$.

Here we design a simple method for the agent to ask advisor initiatively. The action requester enables the agent to learn quickly when to seek advisor assistance. This asking ability of the agent is learnable by traditional reinforcement learning method. It is remarkable that although we add prior advisor knowledge and our trajectory is generated by sampling actions according to two policy, advisor policy and agent policy, our method can still update the network by the normal actor loss and value loss function without additional importance sampling to correct bias. With our action requester method, we can assume that advisor is part of the environment. The action $A_{ask}$, which directly corresponds to a specific standard action, can also be regarded as a reasonable action in the~environment~model.

However, there is still a problem with this method. Although agent can learn to initiatively ask advisor for help, agent may be too dependent on advisor to learn their policy. To solve this problem, we propose a new advisor loss function $\mathcal{L}_{adv}$ according to the supervised learning, in which we treat the advisor action as the prior label information to optimize the parameters of the policy function:
\begin{align}
  \label{eq:adv}
 \mathcal{L}_{adv}(\theta,\phi) = \frac{1}{|\mathcal{D}^n_{adv}|}{ \sum_{(s, a^*) \in \mathcal{D}^n_{adv}}} \Big( \mathcal{L}_{cls}&\big(a^*, \pi_{\theta}(s)\big) +\notag\\& \mathcal{L}_{cls}\big(A_{exec}, g_{\phi}(s)\big) \Big),
\end{align}
where the $\mathcal{L}_{cls}$ is the classification loss function calculated by the traditional cross-entropy, and $\mathcal{D}^n_{adv}$ is the set consisted of the uncertain states with the advisor actions in the $n^{\text{th}}$ iteration. Through this advisor loss function $\mathcal{L}_{adv}$, the agent can learn the correct policy of the uncertain states and reduce its asking demand for these states. Generally, our proposed action requester endows the agent with a novel ability of initially-asking action, allowing the agent to seek advisor assistance when it encounters uncertain states.

\subsection{Adaptive State Selector}

\begin{figure}[!t]
  \centering
  \includegraphics[scale=0.4]{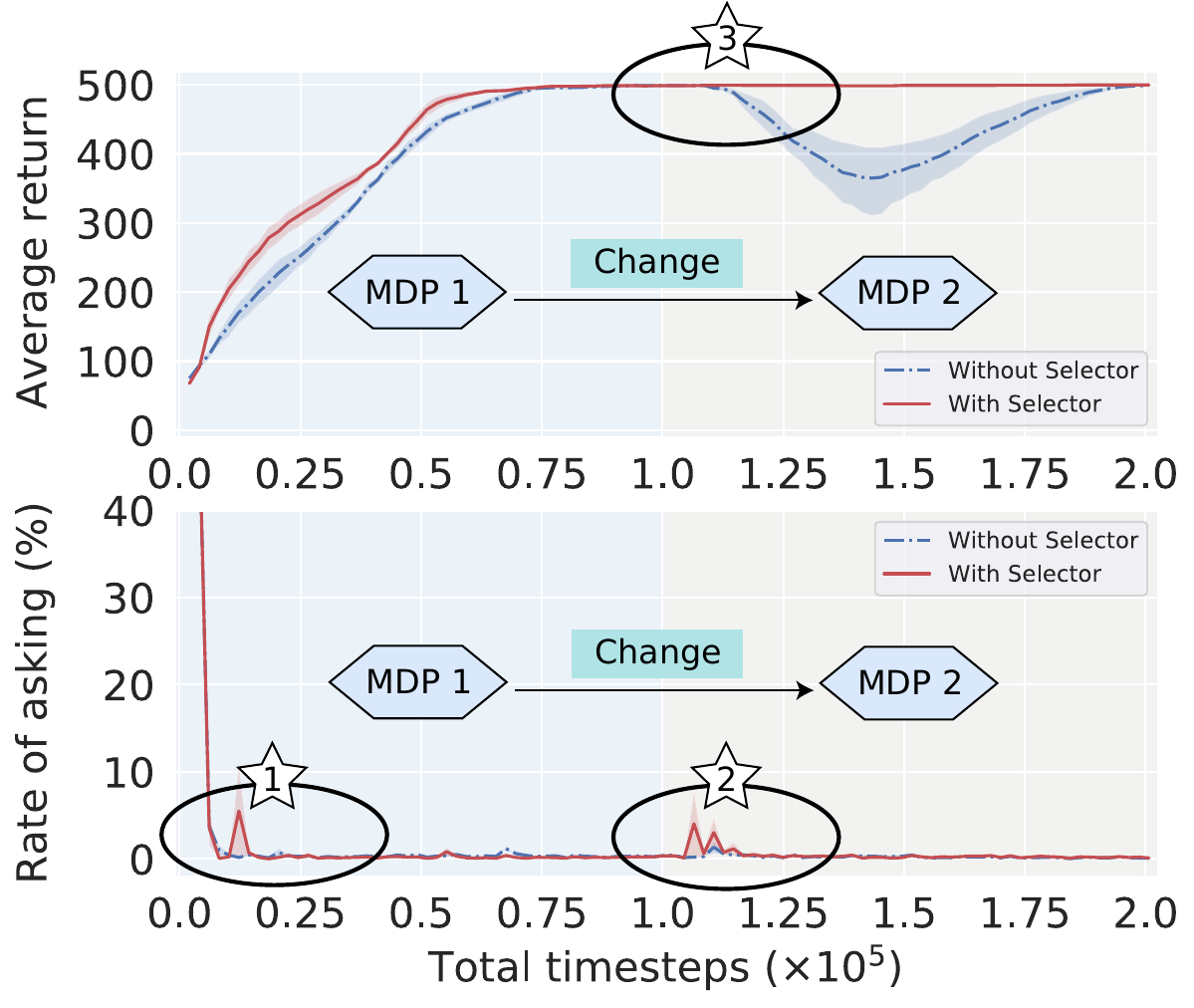}
  \caption{
    Comparing the Ask-AC framework with and without the adaptive state selector in the non-stationary environment, where the environment model changes at the $1.0\times10^5$ step. The shaded region represents one standard deviation of the average evaluation over 5 trials.}
    \vspace{-0.4cm}
  \label{fig:selector}
\end{figure}

Our Ask-AC framework is incomplete if it only possesses the action requester.\footnote{To test the performance of our framework with and without the adaptive state selector, we first integrate the proposed Ask-AC with the Proximal Policy Optimization algorithm~\cite{PPO} as a baseline agent. Then we design a simple non-stationary CartPole environment by modifying the internal parameter to test the robustness of the agent. We initialize the length of the pole $L$ to 0.5, and then modify the length of the pole $L$ to 1.0 at the $1.0\times10^5$ step. Note that we do not tell the agent the modified length parameters.} As shown in Figure~\ref{fig:selector}, if there is only the action requester and no adaptive state selector, the ability of an agent to ask advisor will be gradually inhibited by the advisor loss function as the training goes on~(Star 1). The action requester will gradually miss some unstable states which directly impair the final convergence performance~(Star 3). Especially when the environment changes, the agent without the adaptive state selector fails to perceive the critical states that require advisor feedback~(Star 2). It takes a lot of time for the agent to get back to the original level~(Star 3). To solve this problem, we introduce the adaptive state selector to identify the unstable states potentially missed by the action requester in its history states, and learn to promote the ask action on these states~(Star 1\&2). With the help of the adaptive state selector, our agent can make more effective use of advisor assistance to enhance performance~(Star 3).

The core idea of adaptive state selector is to compare the difference between the current state value estimate $V_{\psi }$ and the return $G_t$. We divide the learning process of the agent into three phases. In the first initial learning phase, the output of value function cannot correctly fit the return. The values of many states are unstable, so the agent needs to seek more advisor feedback. In the second phase, the agent policy and value function have converged. The agent has a certain confidence in the traversed history states, thus the agent no longer or rarely needs to ask advisor for help. In the third phase, when the environment model changes, the difference between the converged output of value function and the return suddenly increases. The states become unstable again. Therefore, the ask action on these states should be promoted according to the current difference. It is worth noting that the third stage may also occur between the first and second phases, but in the same way.

Based on the above ideas, we design the adaptive state selector to detect the unstable state set. Consider the set of history states $\mathcal{S}_{h}^n$ traversed in the $n^{\text{th}}$ iteration, the value loss $\mathcal{L}_{v}^n$ is defined as
\begin{align}
  \mathcal{L}_{v}^n(\psi) = \frac{1}{\left| \mathcal{S}_{h}^n \right|} \sum_{s \in \mathcal{S}_{h}^n} E_v^n(s),
\end{align}
where $E_v^n(s) = \left|\left| V_{\psi}^n(s) - G_t^n(s) \right|\right| ^2$ is the value error between the current state value estimate $V_{\psi}^n(s)$ and the return $G_t^n(s)$ of each state $s$. The exponentially weighted moving average $\mathcal{W}_{v}^n$ of $\mathcal{L}_{v}^n$ over the iterations is updated via
\begin{align}
    \mathcal{W}_{v}^n = \beta * \mathcal{W}_{v}^{n-1} + (1-\beta) * \mathcal{L}_{v}^n,
\end{align}
where the hyper-parameter $\beta$ controls the exponential decay rate of the moving average. Then we define the adaptive unstable rate $R_u^n$ as
\begin{align}
  \label{eq:rate}
    R_u^n = \frac{(1-\beta) * \mathcal{L}_{v}^n}{\mathcal{W}_{v}^n}.
\end{align}
It is easy to show that $R_u^n \in [0,1]$. This adaptive unstable rate mainly focused on the sharp rise of $\mathcal{L}_{v}^n$ over the iterations. If there is a sharp rise of $\mathcal{L}_{v}^n$ in the $n^{\text{th}}$ iteration, the adaptive $R_u^n$ will tend to be a larger value. The number of unstable states $k_u^n$ is calculated by
\begin{align}
  \label{eq:number}
    k_u^n = \big\lceil R_{u}^n * \delta * \left| \mathcal{S}_{h}^n \right| \big\rceil,
\end{align}
where $\delta$ is the max unstable rate. We use $\delta$ to limit the size of the unstable set, which makes agent not depend on advisor too much.

\algdef{SE}[SUBALG]{Indent}{EndIndent}{}{\algorithmicend\ }%
\algtext*{Indent}
\algtext*{EndIndent}
\algnewcommand{\LineComment}[1]{\State \textcolor{gray}{\(\triangleright\) #1}}

\begin{algorithm}[!b]
  \vspace{-0.1cm}
	\caption{Learning Algorithm for Ask-AC}
	\label{alg:learn}
  \begin{flushleft}
    \textbf{Input:} Advisor, action requester $g_{\phi}$, actor network $\pi_\theta$, value network $V_\psi$, exponential decay rate $\beta$, max unstable rate $\delta$ \\
    \textbf{Output:} Optimized network parameters $\phi, \theta, \psi$
  \end{flushleft}
	\begin{algorithmic}[1]
    \State Initialize the network parameters $\phi, \theta, \psi$
    \State Initialize the weighted moving average $\mathcal{W}^0_v = 0$

    \For{each iteration $n = 1 \text{ to } N$}
      \State $e^n \leftarrow$ SampleEpisode(advisor, $g_{\phi}$, $\pi_\theta$) (refer to Alg.~\ref{alg:sample})
      \LineComment{(1) Compute the advisor loss}
      \State Collect $\mathcal{D}^n_{adv}$ from $e^n$
      \State Compute the advisor loss $\mathcal{L}_{adv}(\theta,\phi)$ based on $\mathcal{D}^n_{adv}$
      \LineComment{(2) Compute the ask loss}
      \State Collect $\mathcal{S}_{h}^n$ from $e^n$
      \State Compute the value error $E_v^n(s)$ based on $\mathcal{S}_{h}^n$
      \State Compute the value loss $\mathcal{L}_{v}^n(\psi)$
      \State Update $\mathcal{W}_{v}^n = \beta * \mathcal{W}_{v}^{n-1} + (1-\beta) * \mathcal{L}_{v}^n$
      \State Calculate the adaptive unstable rate $R_u^n$
      \State Calculate the unstable state number $k_u^n$
      \State Sort $\mathcal{S}_{h}^n$ in descending order as $\tilde{\mathcal{S}}_{h}^n$ based on $E_v^n(s)$
      \State Construct $\mathcal{S}_{u}^n$ from the top $k_u^n$ states in $\tilde{\mathcal{S}}_{h}^n$
      \State Compute the ask loss $\mathcal{L}_{ask}(\phi)$ based on $\mathcal{S}_{u}^n$
      \LineComment{(3) Compute the actor-critic loss}
      \State Compute the original method loss $\mathcal{L}_{org}(\theta)$
      \LineComment{(4) Perform gradient descent on the total loss}
      \State Update the network parameters $\phi, \theta, \psi$
    \EndFor
    \State \Return $\phi, \theta, \psi$
	\end{algorithmic}
\end{algorithm}

Finally, $\mathcal{S}_{u}^n$ is the unstable state set consisted of the top $k_u^n$ elements in the set $\tilde{\mathcal{S}}_{h}^n$, where $\tilde{\mathcal{S}}_{h}^n$ is the descending order set of states $s \in \mathcal{S}_{h}^n$ according to $E_v^n(s)$:
\begin{align}
  \label{eq:order}
    \tilde{\mathcal{S}}_{h}^n = \mathop{\arg \operatorname{sort}}\limits_{s \in \mathcal{S}_{h}^n} E_v^n(s),
\end{align}
and then the ask loss function is defined as follow:
\begin{align}
  \label{eq:ask}
 \mathcal{L}_{ask}(\phi) = \frac{1}{|\mathcal{S}_{u}^n|} \sum_{s \in \mathcal{S}_{u}^n} \mathcal{L}_{cls}\big(A_{ask}, g_{\phi}(s)\big),
\end{align}
where the $\mathcal{L}_{cls}$ is calculated by the traditional cross entropy as in the advisor loss function. With this ask loss function, the agent can identify the unstable states and promote the ask action on these states to acquire advisor guidance.

\begin{algorithm}[!t]
	\caption{Sample Episode for Ask-AC}
	\label{alg:sample}
  \begin{flushleft}
    \textbf{Input:} Advisor, action requester $g_{\phi}$, actor network $\pi_\theta$  \\
    \textbf{Output:} Episode $e$
  \end{flushleft}
	\begin{algorithmic}[1]
    \State Obtain the initial state $s_0$
    \While{not terminal}
      \State Make the binary decision $y_t \sim g_{\phi}(s_t)$
      \If{$y_t = A_{ask}$}
        \LineComment{Ask for advisor feedback}
        \State Send the state $s_t$ to the advisor
        \State Receive the action $a_t$ from the advisor
      \ElsIf{$y_t = A_{exec}$}
        \State Sample the action $a_t \sim \pi_{\theta}(s_t)$
      \EndIf
      \State Obtain the reward $r_{t} = \mathcal{R}(s_t, a_t)$
      \State Obtain the next state $s_{t+1} = \mathcal{P}(s_{t+1} | s_t, a_t)$
    \EndWhile
    \State \Return $e = (s_0, y_0, a_0, r_0,\cdots, s_T, y_T, a_T, r_T)$
	\end{algorithmic}
\end{algorithm}

\subsection{Optimization Objective}

Overall, with the action requester and the adaptive state selector, the total loss of our framework is
\begin{align}
  \label{eq:total}
 \mathcal{L}_{total}&(\theta,\psi,\phi) = \notag\\&\mathcal{L}_{org}(\theta,\psi,\phi) + \lambda_{adv} \mathcal{L}_{adv}(\theta,\phi) +  \lambda_{ask} \mathcal{L}_{ask}(\phi),
\end{align}
where the $\mathcal{L}_{org}(\theta,\psi,\phi)$ is the loss of the original method including actor loss and value loss, $\lambda_{adv}$ and $\lambda_{ask}$ are coefficients. An algorithmic description of the training procedure is given in Alg.\ref{alg:learn} and Alg.\ref{alg:sample}. 

\textbf{Complexity Analysis.} The dimensions for all input features and hidden features are assumed $d$ for simplicity.
Let $|\mathcal{A}|$ be the number of actions, and $l$ denotes the number of layers.
The shared embedding layers have $\mathcal{O}\left(ld^2\right)$ parameters. The actor head and critic head have $\mathcal{O}\left(|\mathcal{A}|d + ld^2\right)$ and $\mathcal{O}\left(ld^2\right)$ parameters, respectively. The action requester has $\mathcal{O}\left(ld^2\right)$ parameters. Thus, the total parameters of the Ask-AC is $\mathcal{O}\left(|\mathcal{A}|d + ld^2\right)$. Moreover, the agent needs to sample $M$ transitions at each iteration during training. 
As a result, the space complexity of Ask-AC is $\mathcal{O}\left(Md + |\mathcal{A}|d + ld^2\right)$.
The time complexity of feed-forward propagation is $\mathcal{O}\left(ld^2\right)$, and the time complexity of state sorting is $\mathcal{O}\left(M \log M\right)$.
Overall, the time complexity of Ask-AC is $\mathcal{O}(M \log M + ld^2)$.

\begin{table}[!t] 
  \centering
  \caption{Hyperparameters used in CartPole and LunarLander experiments. $\epsilon$ is linearly annealed from 1 to 0 over the course of learning.}
  \resizebox{0.49\textwidth}{!}{%
  \begin{tabular}{@{}ccc@{}}
      \toprule
                                   & \makecell[c]{PPO / AskPPO / \\ CM / Heu}           & A2C / AskA2C            \\ \midrule
      Policy network hidden layers & {[}64, 64{]}         & {[}64, 64{]}          \\ \specialrule{0em}{0.6pt}{0.6pt}
      Value network hidden layers  & {[}64, 64{]}         & {[}64, 64{]}          \\ \specialrule{0em}{0.6pt}{0.6pt}
      Timesteps per iteration      & 2048                 & 40                    \\ \specialrule{0em}{0.6pt}{0.6pt}
      Learning rate                & 0.001$\times \epsilon$ & 0.0007$\times \epsilon$ \\ \specialrule{0em}{0.6pt}{0.6pt}
      Number of epochs             & 10                   & N/A                   \\ \specialrule{0em}{0.6pt}{0.6pt}
      Minibatch size               & 256                  & N/A                   \\ \specialrule{0em}{0.6pt}{0.6pt}
      Discount factor              & 0.99                 & 0.99                  \\ \specialrule{0em}{0.6pt}{0.6pt}
      GAE discount                 & 0.95                 & 1.0                   \\ \specialrule{0em}{0.6pt}{0.6pt}
      PPO clipping                 & 0.2$\times \epsilon$ & N/A                   \\ \specialrule{0em}{0.6pt}{0.6pt}
      Gradient clipping            & 0.5                  & 0.5                   \\ \specialrule{0em}{0.6pt}{0.6pt}
      VF coeff                     & 0.5                  & 0.5                   \\ \specialrule{0em}{0.6pt}{0.6pt}
      Entropy coeff                & 0.0                  & 0.0                   \\  \specialrule{0em}{0.6pt}{0.6pt}
      Exponential decay rate (Ask) & 0.9                  & 0.9                   \\  \specialrule{0em}{0.6pt}{0.6pt}
      Max unstable rate (Ask)      & 0.1                  & 0.1                   \\  \specialrule{0em}{0.6pt}{0.6pt}
      Advisor loss coeff (Ask)       & 1.0                  & 1.0                   \\  \specialrule{0em}{0.6pt}{0.6pt}
      Ask loss coeff (Ask)         & 0.5                  & 0.5                   \\  \bottomrule
      \end{tabular}}
  \label{tab:CartPole}
\end{table}

\begin{table}[!t]
  \centering
  \caption{Hyperparameters used in DoorKey and MultiRoom experiments.}
  \resizebox{0.49\textwidth}{!}{%
  \begin{tabular}{@{}ccc@{}}
  \toprule
                               & \makecell[c]{PPO / AskPPO / \\ CM / Heu}   & A2C / AskA2C   \\ \midrule
  Policy network hidden layers & {[}64, 64{]} & {[}64, 64{]} \\  \specialrule{0em}{0.6pt}{0.6pt}
  Value network hidden layers  & {[}64, 64{]} & {[}64, 64{]} \\ \specialrule{0em}{0.6pt}{0.6pt}
  Timesteps per iteration      & 1024         & 40           \\ \specialrule{0em}{0.6pt}{0.6pt}
  Learning rate                & 0.00025       & 0.0007         \\ \specialrule{0em}{0.6pt}{0.6pt}
  Number of epochs             & 10           & N/A          \\ \specialrule{0em}{0.6pt}{0.6pt}
  Minibatch size               & 64           & N/A          \\ \specialrule{0em}{0.6pt}{0.6pt}
  Discount factor              & 0.99         & 0.99         \\ \specialrule{0em}{0.6pt}{0.6pt}
  GAE discount                 & 0.95         & 1.0          \\ \specialrule{0em}{0.6pt}{0.6pt}
  PPO clipping                 & 0.2          & N/A          \\ \specialrule{0em}{0.6pt}{0.6pt}
  Gradient clipping            & 0.5          & 0.5          \\ \specialrule{0em}{0.6pt}{0.6pt}
  VF coeff                     & 0.5          & 0.5          \\ \specialrule{0em}{0.6pt}{0.6pt}
  Entropy coeff                & 0.0          & 0.0          \\ \specialrule{0em}{0.6pt}{0.6pt}
  Exponential decay rate (Ask) & 0.9          & 0.9           \\ \specialrule{0em}{0.6pt}{0.6pt}
  Max unstable rate (Ask)      & 0.1          & 0.1           \\   \specialrule{0em}{0.6pt}{0.6pt}
  Advisor loss coeff (Ask)       & 1.0          & 1.0          \\  \specialrule{0em}{0.6pt}{0.6pt}
  Ask loss coeff (Ask)         & 0.5          & 0.5          \\  \bottomrule
  \end{tabular}}
  \label{tab:DoorKey}
\end{table}

\section{Experiments}

To demonstrate the effectiveness of the proposed Ask-AC framework, experiments are conducted based on the stationary environments: \emph{LunarLander-v2} and \emph{CartPole-v1}~\cite{gym}, \emph{MultiRoom-N2-S4-v0} and \emph{DoorKey-v0}~\cite{gym_minigrid} and the non-stationary environments: \emph{Non-stationary CartPole} and \emph{Non-stationary DoorKey}, which are modified from the original environments. In this section, we first introduce the compared methods and the detailed parameter settings. Moreover, the interaction process and the metric are elaborated on. Then the comparison results and robustness analysis are reported to evaluate the performance of the proposed framework\footnote[1]{Our Code is available at \url{https://github.com/liushunyu/Ask-AC}.}.

\subsection{Experimental Settings}

\paragraph{Comparison Method and Parameter} 
\textbf{(1)~Original method:} the Advantage Actor Critic algorithm~\cite{A3C} and Proximal Policy Optimization algorithm~\cite{PPO}, termed as \emph{A2C} and \emph{PPO}. 
\textbf{(2)~Our initiative method:} we integrate the proposed Ask-AC framework with both the A2C algorithm and PPO algorithm, termed as \emph{AskPPO} and \emph{AskA2C}. 
\textbf{(3)~Continuous-monitoring method:} since the experimental results are sensitive to the hand-crafted rules that are hard to design, we tend to compare our framework with the continuous-monitoring method. However, the architecture and feedback type of existing methods are not consistent with our framework. To make a fair comparison, we directly combine the PPO algorithm with supervised loss using continuous advisor label information, termed as \emph{CM}.
\textbf{(4)~Heuristic method:} Considering the action advising method of computing the difference between maximum and minimum Q value as state importance~\cite{DBLP:conf/ijcai/AmirKKG16}, we design a similar heuristic method based on the PPO algorithm using action probability to compute the state importance $I(s) = \max_a{\pi_{\theta}(a|s)} - \min_a{\pi_{\theta}(a|s)}$, termed as \emph{Heu}. When the state importance $I(s)$ is below the pre-defined threshold $\sigma$, the agent can ask advisor for action feedback. The specific experimental parameters are given in Table~\ref{tab:CartPole} and Table~\ref{tab:DoorKey}, where the training parameters of our method and the original one are consistent except for those in the advisor-in-the-loop framework to ensure comparability. The coefficients of advisor loss and ask loss are set as $\lambda_{adv}=1.0$ and $\lambda_{ask}=0.5$, the exponential decay rate and the max unstable rate are set as $\beta=0.9$ and $\delta=0.1$.

\begin{figure*}[!b]
  \centering
  \subfloat[LunarLander]{\includegraphics[width=0.25\textwidth]{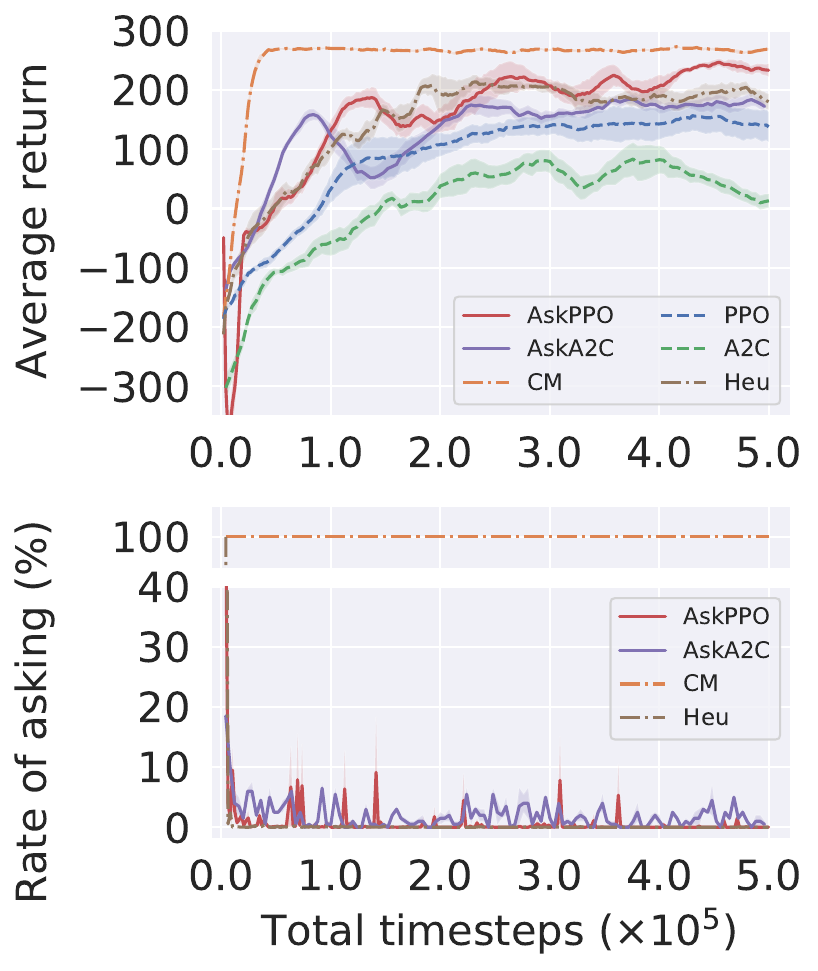}}
  \subfloat[CartPole ($L=0.5$)]{\includegraphics[width=0.25\textwidth]{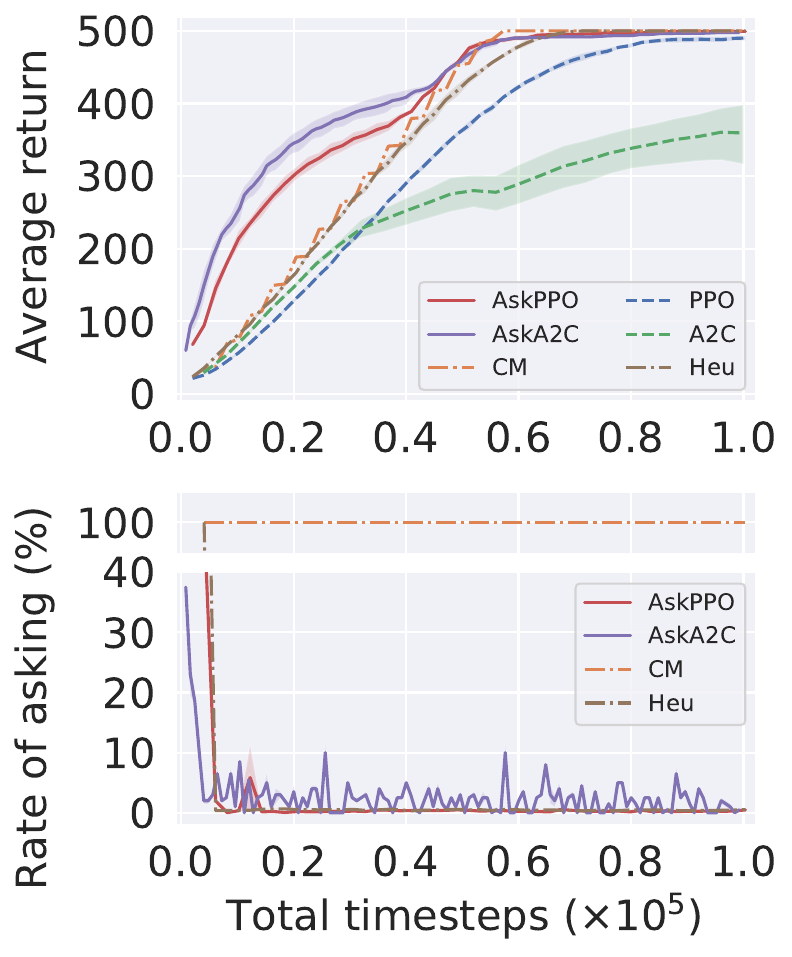}}
  \subfloat[CartPole ($L=1.0$)]{\includegraphics[width=0.25\textwidth]{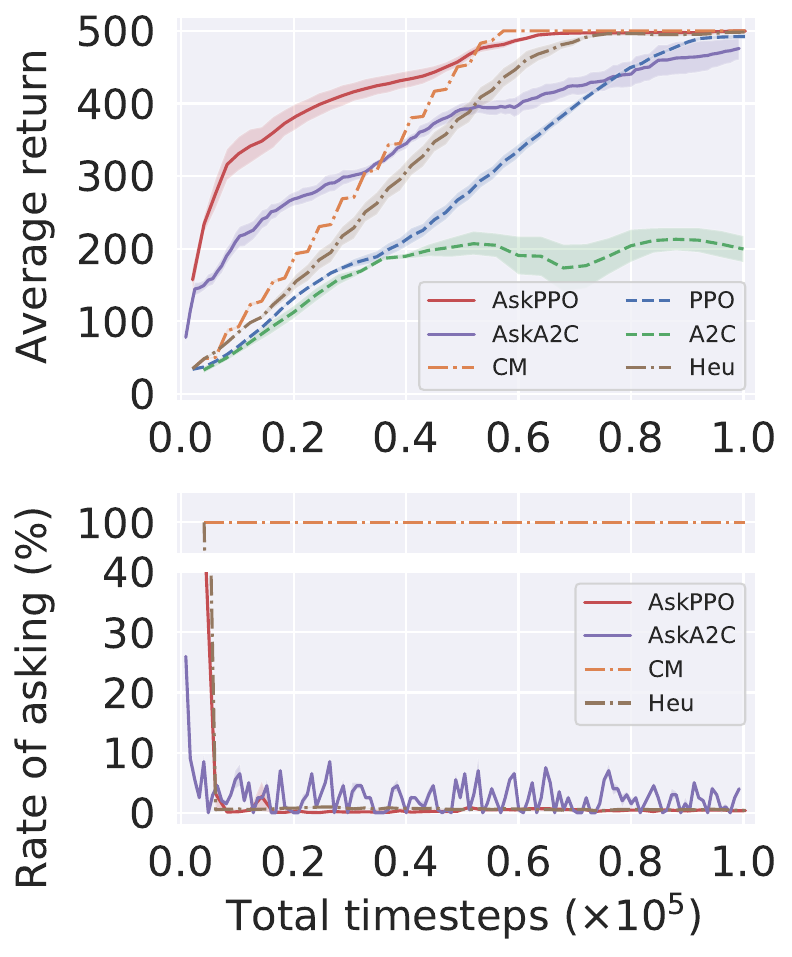}}
  \subfloat[CartPole ($L=2.0$)]{\includegraphics[width=0.25\textwidth]{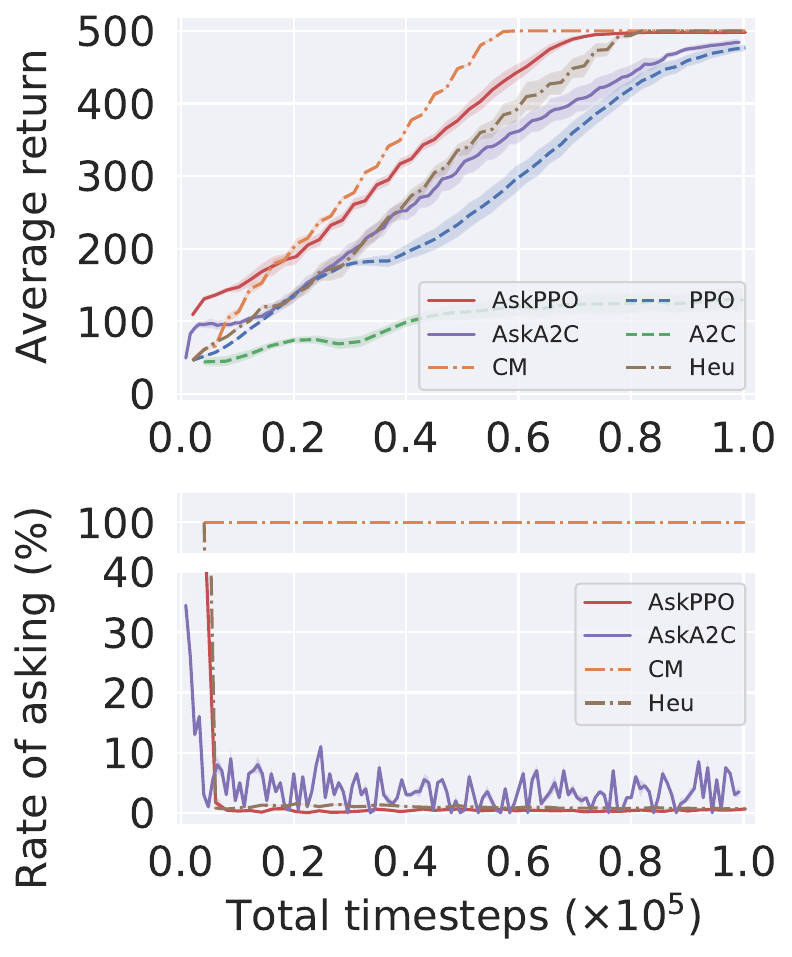}}

  \subfloat[MultiRoom]{\includegraphics[width=0.25\textwidth]{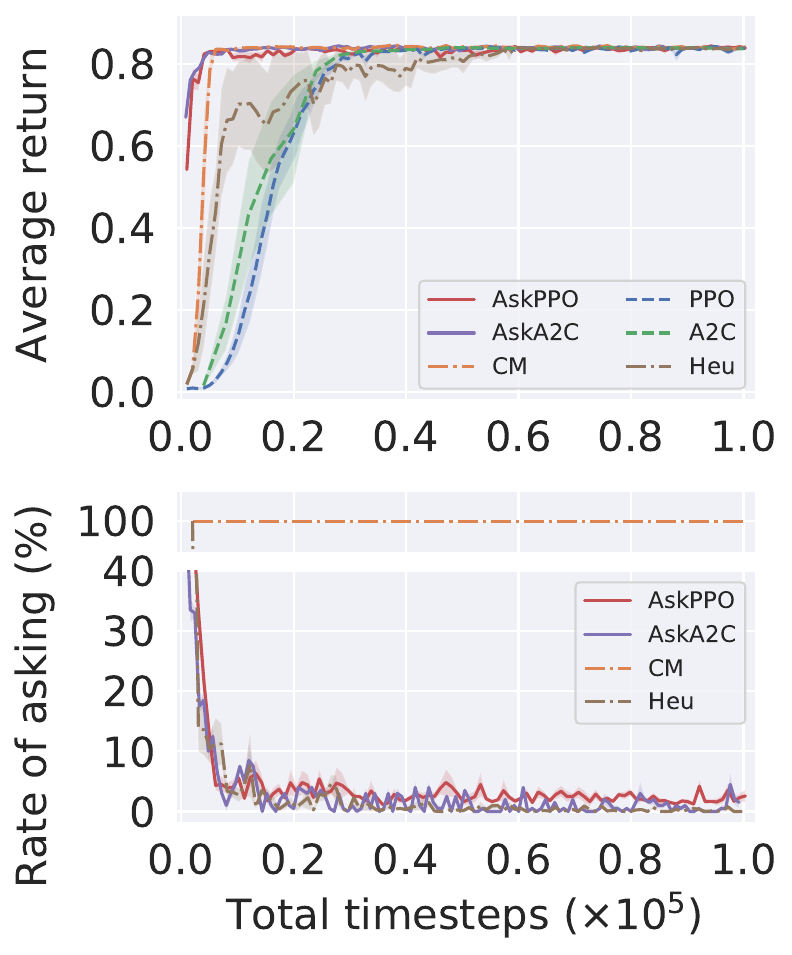}}
  \subfloat[DoorKey ($S=5\times5$)]{\includegraphics[width=0.25\textwidth]{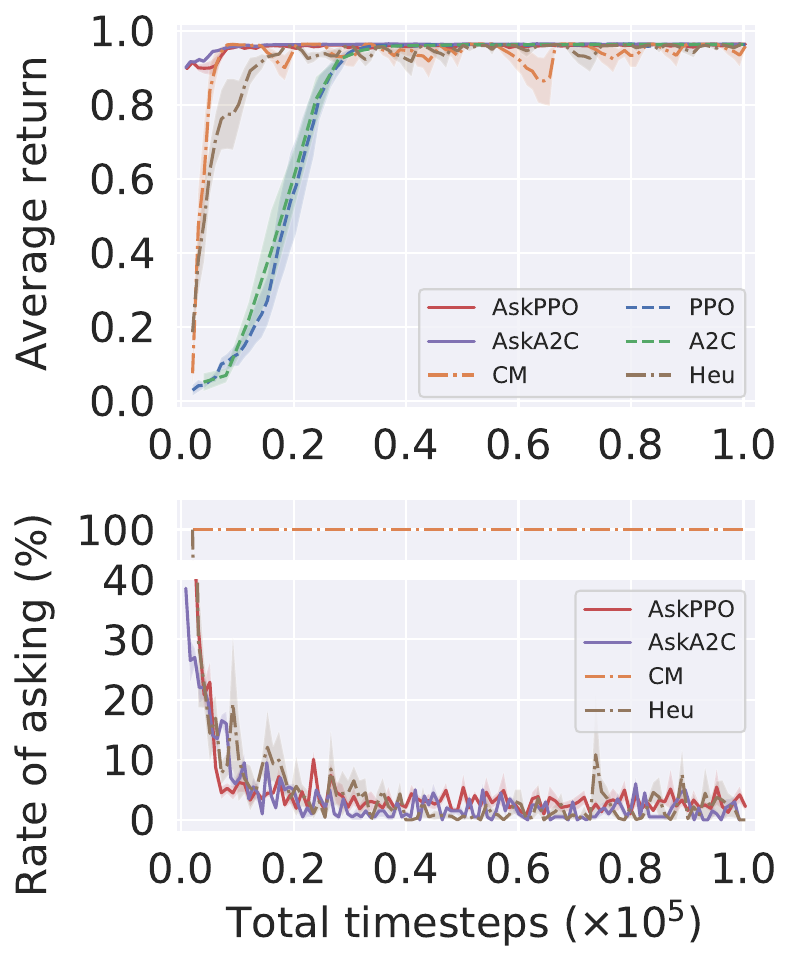}}
  \subfloat[DoorKey ($S=6\times6$)]{\includegraphics[width=0.25\textwidth]{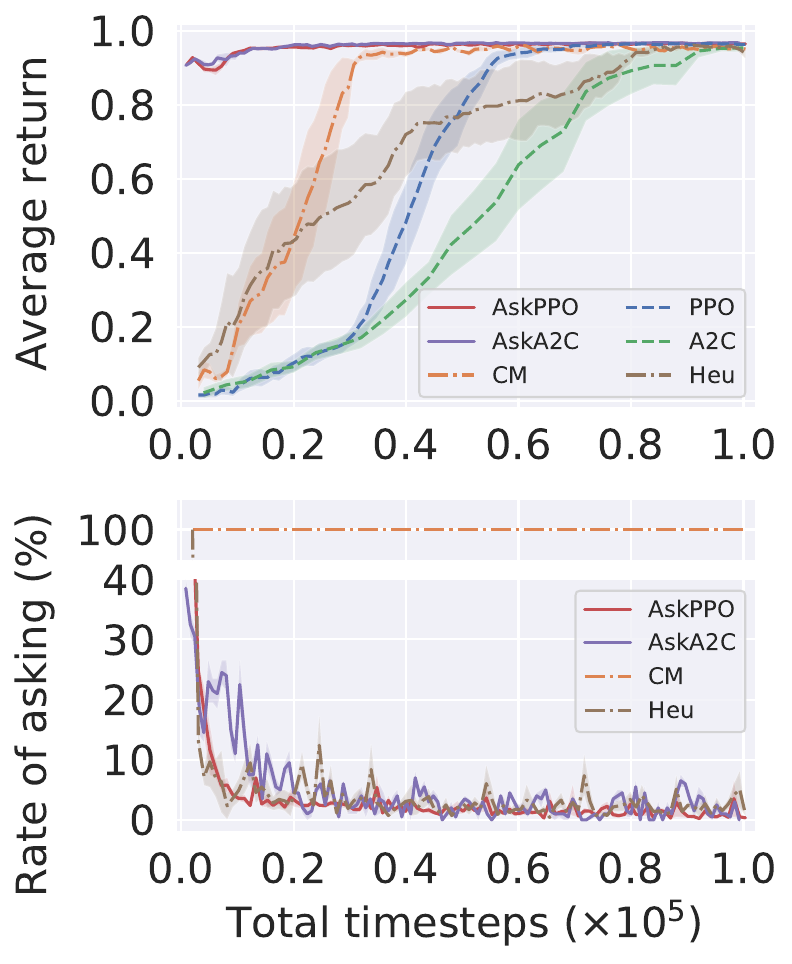}}
  \subfloat[DoorKey ($S=8\times8$)]{\includegraphics[width=0.25\textwidth]{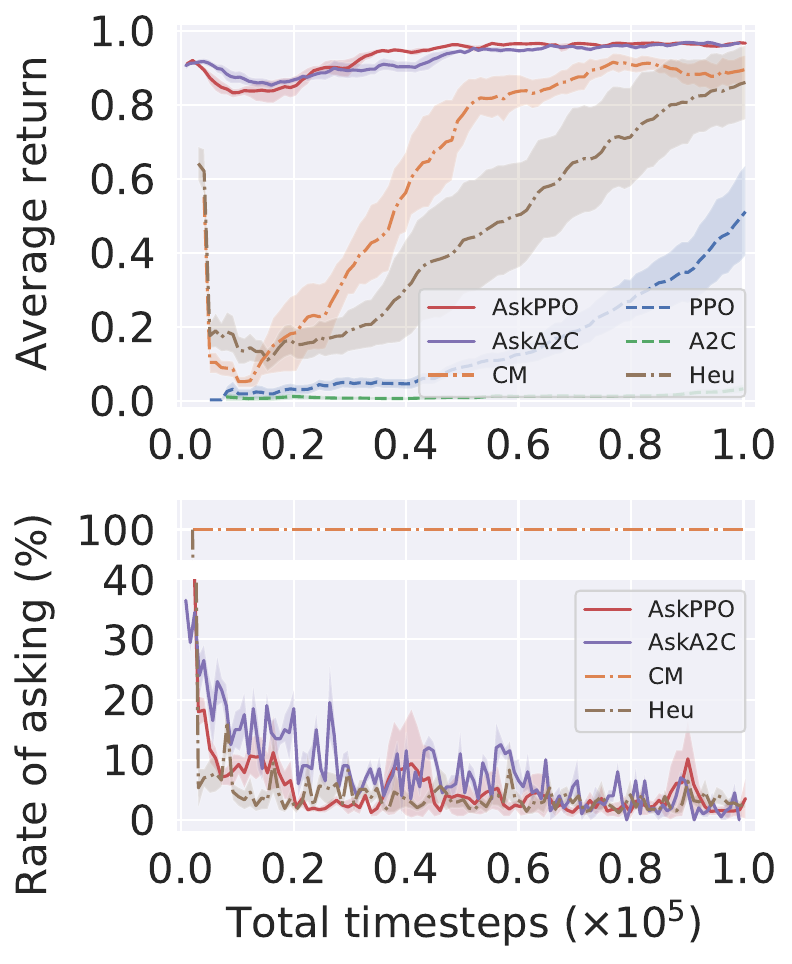}}
  \vspace{0.2cm}
  \caption{Learning curves for eight tasks in four stationary environments. We change the CartPole and DoorKey into six tasks by modifying the internal parameters to test the performance of our method over varying difficulty. For CartPole, we modify the length of the pole $L$ to 0.5, 1.0 and 2.0. For the DoorKey, we modify the size of the maze $S$ to $5\times5$, $6\times6$ and $8\times8$. The shaded region represents one standard deviation of the average evaluation over 5 trials.}
  \label{fig:stationary}
\end{figure*}

\begin{table*}[!t]
  \centering
  \caption{
  The performance comparison between the Ask-AC framework and other methods for eight tasks in four stationary environments. AR is obtained in the testing process and $\pm$ corresponds to one standard deviation of the average evaluation over 10 trials, while ANR and SER are obtained in the training process. Note that the lower the value of ANR and SER, the higher the learning efficiency of the method. \textbf{Bold} indicates that our method achieves close results on par with the CM method. {\color[HTML]{CB0000} Red} indicates that our method reduces the ANR compared with the CM method and heuristic method while our method achieves the better SER than the original method and heuristic method.}
  \vspace{0.2cm}
  \label{tab:stationary-1}
  \resizebox{\textwidth}{!}{%
  \begin{tabular}{@{}ccccccccccccc@{}}
    \toprule
                                   & \multicolumn{3}{c}{LunarLandar}                                                                             & \multicolumn{3}{c}{CartPole ($L=0.5$)}                                                                        & \multicolumn{3}{c}{CartPole ($L=1.0$)}                                                   & \multicolumn{3}{c}{CartPole ($L=2.0$)}                                                   \\ \cmidrule(l){2-13} 
                                   & \cellcolor[HTML]{EFEFEF}AR & \cellcolor[HTML]{EFEFEF}ANR & \multicolumn{1}{c|}{\cellcolor[HTML]{EFEFEF}SER} & \cellcolor[HTML]{EFEFEF}AR & \cellcolor[HTML]{EFEFEF}ANR & \multicolumn{1}{c|}{\cellcolor[HTML]{EFEFEF}SER} & \cellcolor[HTML]{EFEFEF}AR & \cellcolor[HTML]{EFEFEF}ANR & \cellcolor[HTML]{EFEFEF}SER & \cellcolor[HTML]{EFEFEF}AR & \cellcolor[HTML]{EFEFEF}ANR & \cellcolor[HTML]{EFEFEF}SER \\ \midrule
    A2C                            & 94 $\pm$ 95               & -                           & 100\%                                            & 474 $\pm$ 79               & -                           & 100\%                                            & 221 $\pm$ 92               & -                           & 100\%                       & 187 $\pm$ 66               & -                           & 100\%                       \\ \specialrule{0em}{1pt}{1pt}
    \cellcolor[HTML]{FFCCC9}AskA2C & \textbf{223 $\pm$ 75}      & {\color[HTML]{CB0000} 9\%}  & {\color[HTML]{CB0000} 14\%}                      & \textbf{500 $\pm$ 0}       & {\color[HTML]{CB0000} 4\%}  & {\color[HTML]{CB0000} 18\%}                      & \textbf{500 $\pm$ 0}      & {\color[HTML]{CB0000} 5\%}  & {\color[HTML]{CB0000} 11\%} & \textbf{500 $\pm$ 0}      & {\color[HTML]{CB0000} 12\%}  & {\color[HTML]{CB0000} 17\%} \\ \midrule
    PPO                            & 264 $\pm$ 17               & -                           & 100\%                                            & 500 $\pm$ 0                & -                           & 100\%                                            & 500 $\pm$ 0                & -                           & 100\%                       & 500 $\pm$ 0                & -                           & 100\%                       \\ \specialrule{0em}{1pt}{1pt}
    CM                             & 279 $\pm$ 14               & 100\%                       & 6\%                                              & 500 $\pm$ 0                & 100\%                       & 49\%                                             & 500 $\pm$ 0                & 100\%                       & 49\%                        & 500 $\pm$ 0                & 100\%                       & 49\%                        \\ \specialrule{0em}{1pt}{1pt}
    Heu                     & 255 $\pm$ 47               & 18\%                         & 33\%                                             & 500 $\pm$ 0                & 9\%                         & 53\%                                             & 500 $\pm$ 0                & 9\%                         & 59\%                        & 500 $\pm$ 0                & 10\%                         & 69\%                        \\ \specialrule{0em}{1pt}{1pt}
    \cellcolor[HTML]{FFCCC9}AskPPO & \textbf{274 $\pm$ 18}       & {\color[HTML]{CB0000} 14\%}  & {\color[HTML]{CB0000} 24\%}                      & \textbf{500 $\pm$ 0}       & {\color[HTML]{CB0000} 5\%}  & {\color[HTML]{CB0000} 47\%}                      & \textbf{500 $\pm$ 0}       & {\color[HTML]{CB0000} 5\%}  & {\color[HTML]{CB0000} 45\%} & \textbf{500 $\pm$ 0}       & {\color[HTML]{CB0000} 5\%}  & {\color[HTML]{CB0000} 57\%} \\ \bottomrule \specialrule{0em}{1pt}{1pt}
                                   & \multicolumn{3}{c}{MultiRoom}                                                                               & \multicolumn{3}{c}{\cellcolor[HTML]{FFFFFF}DoorKey ($S=5\times5$)}                                                 & \multicolumn{3}{c}{DoorKey ($S=6\times6$)}                                                    & \multicolumn{3}{c}{DoorKey ($S=8\times8$)}                                                    \\ \cmidrule(l){2-13} 
    & \cellcolor[HTML]{EFEFEF}AR & \cellcolor[HTML]{EFEFEF}ANR & \multicolumn{1}{c|}{\cellcolor[HTML]{EFEFEF}SER} & \cellcolor[HTML]{EFEFEF}AR & \cellcolor[HTML]{EFEFEF}ANR & \multicolumn{1}{c|}{\cellcolor[HTML]{EFEFEF}SER} & \cellcolor[HTML]{EFEFEF}AR & \cellcolor[HTML]{EFEFEF}ANR & \cellcolor[HTML]{EFEFEF}SER & \cellcolor[HTML]{EFEFEF}AR & \cellcolor[HTML]{EFEFEF}ANR & \cellcolor[HTML]{EFEFEF}SER \\ \midrule
    A2C                            & 0.83 $\pm$ 0.04            & -                           & 100\%                                            & 0.96 $\pm$ 0.01            & -                           & 100\%                                            & 0.90 $\pm$ 0.15            & -                           & 100\%                       & 0.12 $\pm$ 0.25            & -                           & 100\%                       \\ \specialrule{0em}{1pt}{1pt}
    \cellcolor[HTML]{FFCCC9}AskA2C & \textbf{0.85 $\pm$ 0.04}   & {\color[HTML]{CB0000} 18\%}  & {\color[HTML]{CB0000} 13\%}                      & \textbf{0.96 $\pm$ 0.01}   & {\color[HTML]{CB0000} 21\%}  & {\color[HTML]{CB0000} 10\%}                      & \textbf{0.96 $\pm$ 0.01}   & {\color[HTML]{CB0000} 8\%}  & {\color[HTML]{CB0000} 10\%} & \textbf{0.96 $\pm$ 0.01}   & {\color[HTML]{CB0000} 24\%}  & {\color[HTML]{CB0000} 10\%} \\ \midrule
    PPO                            & 0.82 $\pm$ 0.05            & -                           & 100\%                                            & 0.96 $\pm$ 0.01            & -                           & 100\%                                            & 0.94 $\pm$ 0.04            & -                           & 100\%                       & 0.80 $\pm$ 0.23            & -                           & 100\%                       \\ \specialrule{0em}{1pt}{1pt}
    CM                             & 0.84 $\pm$ 0.04            & 100\%                       & 11\%                                             & 0.96 $\pm$ 0.01            & 100\%                       & 13\%                                             & 0.95 $\pm$ 0.03            & 100\%                       & 33\%                        & 0.96 $\pm$ 0.03            & 100\%                       & 37\%                        \\ \specialrule{0em}{1pt}{1pt}
    Heu                     & 0.82 $\pm$ 0.05            & 31\%                        & 23\%                                             & 0.96 $\pm$ 0.01            & 37\%                        & 15\%                                             & 0.95 $\pm$ 0.04            & 15\%                         & 78\%                        & 0.90 $\pm$ 0.12            & 12\%                         & 54\%                        \\ \specialrule{0em}{1pt}{1pt}
    \cellcolor[HTML]{FFCCC9}AskPPO & \textbf{0.83 $\pm$ 0.04}   & {\color[HTML]{CB0000} 20\%}  & {\color[HTML]{CB0000} 11\%}                      & \textbf{0.96 $\pm$ 0.01}   & {\color[HTML]{CB0000} 22\%}  & {\color[HTML]{CB0000} 13\%}                      & \textbf{0.96 $\pm$ 0.01}   & {\color[HTML]{CB0000} 7\%}  & {\color[HTML]{CB0000} 11\%} & \textbf{0.97 $\pm$ 0.01}   & {\color[HTML]{CB0000} 6\%}  & {\color[HTML]{CB0000} 10\%} \\ \bottomrule
    \end{tabular}}
  \end{table*}

\paragraph{Interaction Process} 
We mainly study the framework for training in simulation environments. In the setting of reinforcement learning, the agent interacts with the environment step by step. When the agent seeks advisor assistance at a certain step, the simulation environment will present the current state to the advisor expert. Then the advisor expert will select an action as input from the action set. After receiving the feedback, the agent will take the advisor action and interact with the simulation environment again. To carry out sufficient experiments to verify the effectiveness of our framework, we use a trained PPO as an advisor in the stationary environment. For the non-stationary environment, we use multiple trained PPO for different conditions. In our experiment, multiple trained PPO are considered as a single advisor from the perspective of the agent. We assume that the advisor can know the current environment an agent encounters from a given state. Thus, we directly use the corresponding trained PPO to provide action feedback when the agent seeks assistance. In practice, human advisors can also follow the same process to participate in the interaction process.
Moreover, to simulate advisor operation more realistically, we conduct robustness analysis to explore the impact of advisers with different operation levels on the performance of our framework.

\paragraph{Metric}
The metrics we adopted include Average Return~(AR), Rate of Asking~(RoA), Sampling Efficiency Ratio~(SER) and Asking Number Ratio~(ANR). SER is the total timestep ratio between the interactive method~(CM / Heu / AskPPO / AskA2C) and its original method~(PPO / A2C) achieving the consistent highest AR. ANR is the total asking number ratio between the initiative method~(AskPPO / AskA2C) or the heuristic method (Heu) and the continuous-monitoring method~(CM) achieving the foregoing AR. The detailed formulation of the metrics is described as follows:

\begin{itemize}
    \item \emph{Average Return~(AR).} The AR shown in the figure is the performance score of an agent in training process with advisor assistance, and the AR shown in the table is the performance score of an agent in testing process without advisor assistance.
    \item \emph{Rate of Asking~(RoA).} The RoA shown in the figure is the rate of an agent asking advisor in each iteration.
    \item \emph{Sampling Efficiency Ratio~(SER).} If it takes the original agent $T_{org}$ timesteps to achieve its highest average return $AR_{org}$ in training process, and interactive agent $T_{inter}$ timesteps to achieve the consistent average return $AR_{org}$, we define SER to be:
    \begin{align}
        SER = \frac{T_{inter}}{T_{org}}.
    \end{align}
    \item \emph{Asking Number Ratio~(ANR).} If initiative agent or heuristic agent ask advisor $T_{ask}$ times before it achieve the consistent average return $AR_{org}$ mentioned above, and the continuous-monitoring agent ask advisor $T_{cm}$ times, we define ANR to be:
    \begin{align}
        ANR = \frac{T_{ask}}{T_{cm}}.
    \end{align}
\end{itemize}

Note that the interactive agent can achieve the consistent average return with the original agent in our experiment.

\begin{figure*}[!t]
  \centering
  \subfloat[Non-stationary CartPole]{
      \label{Non-stationary CartPole}
      \includegraphics[width=0.49\textwidth]{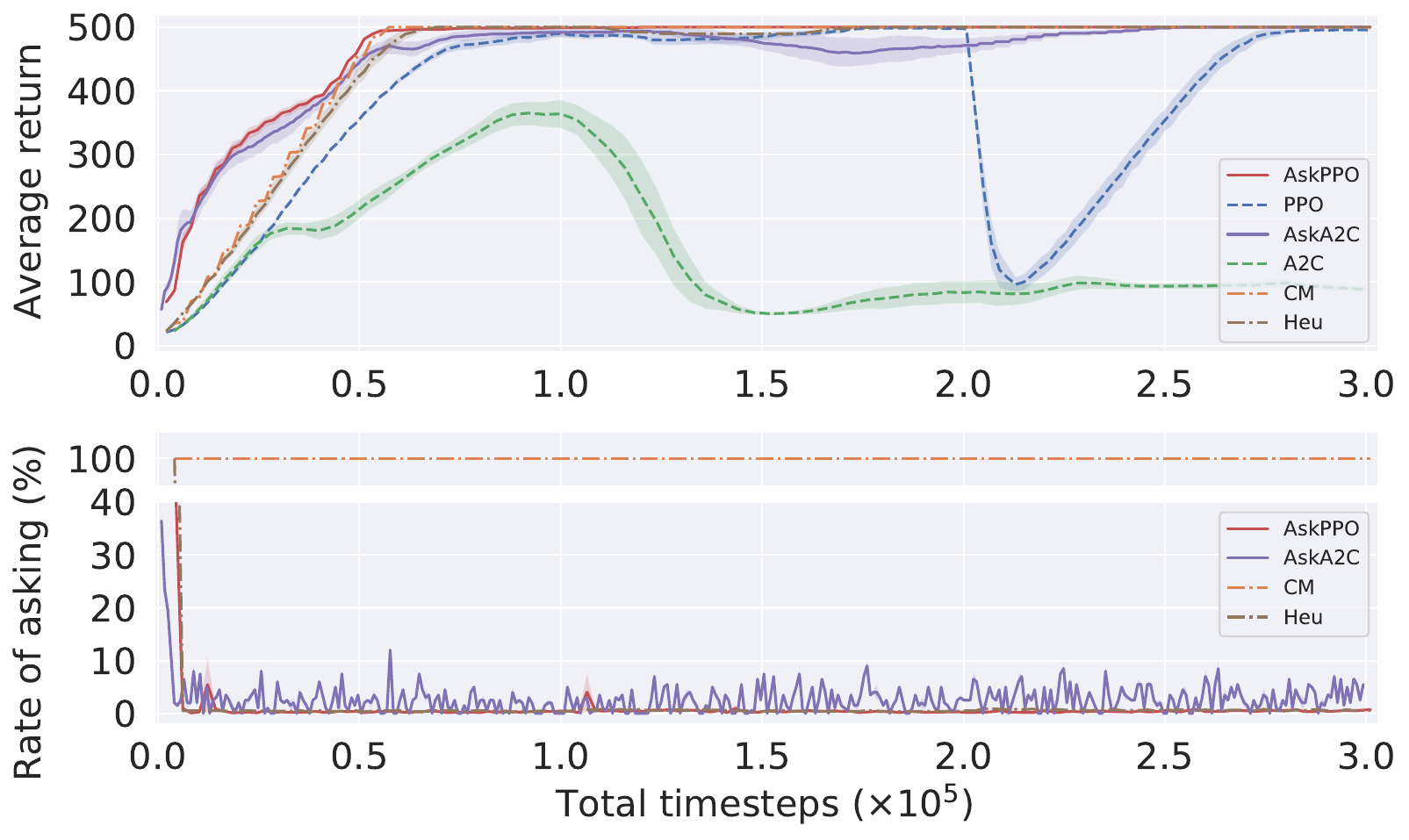}}
  \subfloat[Non-stationary DoorKey]{
      \label{Non-stationary DoorKey}
      \includegraphics[width=0.49\textwidth]{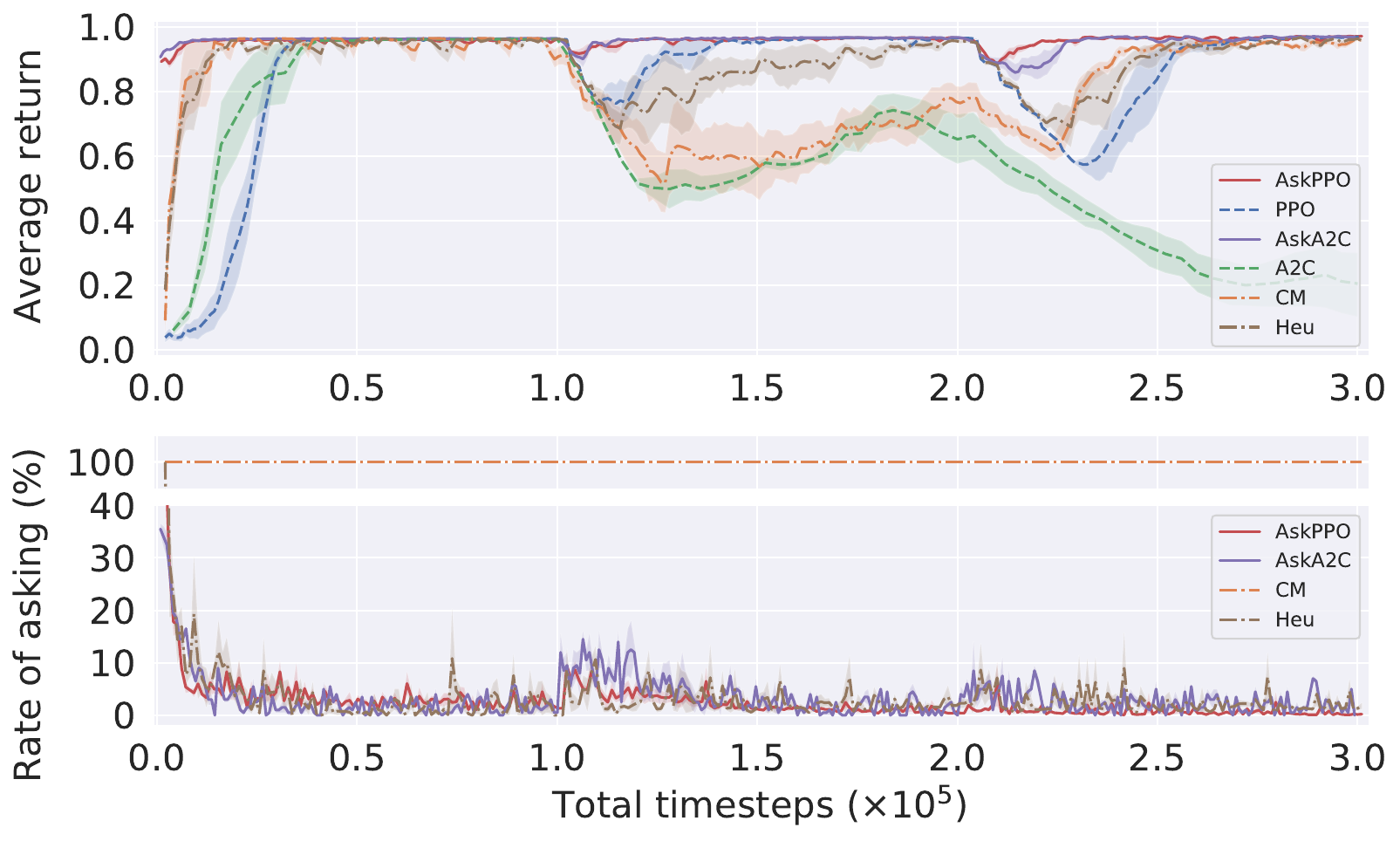}}
      \vspace{0.2cm}
  \caption{Learning curves for the Non-stationary CartPole environment and the Non-stationary DoorKey environment, trained for three hundred thousand timesteps. In the Non-stationary CartPole environment, we initialize the length of the pole $L$ to $0.5$, and then modify the length of the pole $L$ to $1.0$ and $2.0$ at the $1.0\times 10^5$ and $2.0\times 10^5$ step, respectively. In the Non-stationary DoorKey environment, we initialize the size of the maze $S$ to 5x5, and then modify the size of the maze $S$ to $6\times6$ and $8\times8$ at the $1.0\times 10^5$ and $2.0\times 10^5$ step, respectively.}
  \label{fig:non-stationary}
\end{figure*}

\subsection{Stationary Environment}

The experimental results on stationary environment are shown in Figure~\ref{fig:stationary} and Table~\ref{tab:stationary-1}. Compared with the CM method and the heuristic method, our proposed Ask-AC framework reduces the demand for advisor attention~(see ANR) while maintaining high sampling efficiency~(see SER). It is obvious that not all states are worth asking advisor. The agent has its own ability to learn effective policy for simple states, while it only needs to seek advisor assistance when it encounters uncertain states. Furthermore, we remove the action requester in the test experiment to verify that the trained agents master the decision-making ability rather than learning to rely on asking advisor for answers. The experimental results show that our method improves the average return compared with the original method and achieves close results on par with the continuous-monitoring method (see AR). 

Moreover, in several complex environments, the CM method has lower learning efficiency than our proposed Ask-AC. We can explain this phenomenon from the perspective of exploration which serves as a critical factor in reinforcement learning. Since continuous supervisions impose a strong constraint on agent learning, the CM method can be regarded as traditional behavior cloning where the agent learns the policy by supervised learning only based on optimal expert trajectories without any exploration. In recent years, many studies have shown that behavior cloning is susceptible to overfitting and often yields limit effect~\cite{garg2021iq,DBLP:conf/nips/OrsiniRHVDGGBPA21,DBLP:conf/iclr/Ciosek22}. The agent only learns to follow the optimal trajectory. Thus, minor errors can quickly compound when the learned policy departs from the states in the optimal trajectory. In such cases, the agent often performs poorly when tested in the environment. To sum up, the reason why CM agents cannot achieve comparable performance with us is likely due to the risk of overfitting, which causes the problem of insufficient exploration and directly damages the learning efficiency of the agents. On the contrary, Ask-AC enables the agent to explore the environment and initiatively seek advisor intervention in the presence of uncertain states. Therefore, Ask-AC can yield better performance than the CM method in learning efficiency.

Averaging the results over all tasks show that if the original method needs \emph{100} timesteps to achieve the highest return, the heuristic method needs \emph{38.6} timesteps. However, our Ask-AC framework only needs \emph{20.1} timesteps, which achieve the comparable performance with \emph{24.4} timesteps of the CM method. Furthermore, the CM method needs to ask for advisor feedback \emph{100} times and the heuristic method needs \emph{21.7} times, while our Ask-AC framework significantly reduces the demand for advisor attention and only needs to ask \emph{11.5} times.

\subsection{Non-stationary Environment}

The purpose of the non-stationary environment design is to test the robustness of the method, where the goal of the agent is to learn to adjust its policy and rapidly adapt to the environmental change, rather than finding an optimal policy to master all environments. Figure~\ref{fig:non-stationary} shows that our agents can quickly perceive the change and learn to promote the ask action in the non-stationary environment. Then they can rapidly adjust their policies with advisor assistance to perform better than the heuristic agents and CM agents. Furthermore, as shown in Figure~\ref{Non-stationary CartPole}, advisor assistance can improve the generalization ability of the agents to maintain good performance when the environment changes. By contrast, the original agents often lack adaptivity to the change and get a lower return. Nevertheless, the performance of CM agents in Figure~\ref{Non-stationary DoorKey} shows that too much advisor intervention also inevitably limits the exploration of the agent and further damage its generalization ability in the non-stationary environment, resulting in poor performance. In contrast, our Ask-AC agent can learn and generalize the policy in different conditions.

\subsection{Robustness Analysis}

\begin{figure}[!t]
  \centering
  \vspace{-0.3cm}
    \subfloat[CartPole ($L=0.5$)]{\includegraphics[scale=0.32]{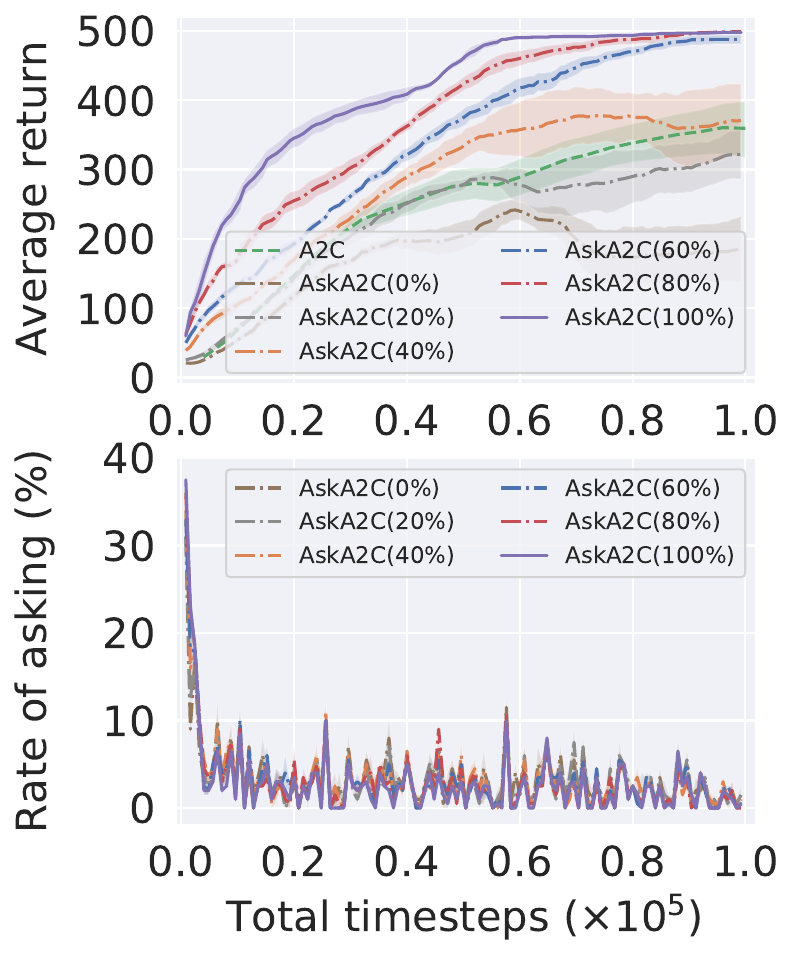}}\;
    \subfloat[DoorKey ($S=5\times5$)]{\includegraphics[scale=0.32]{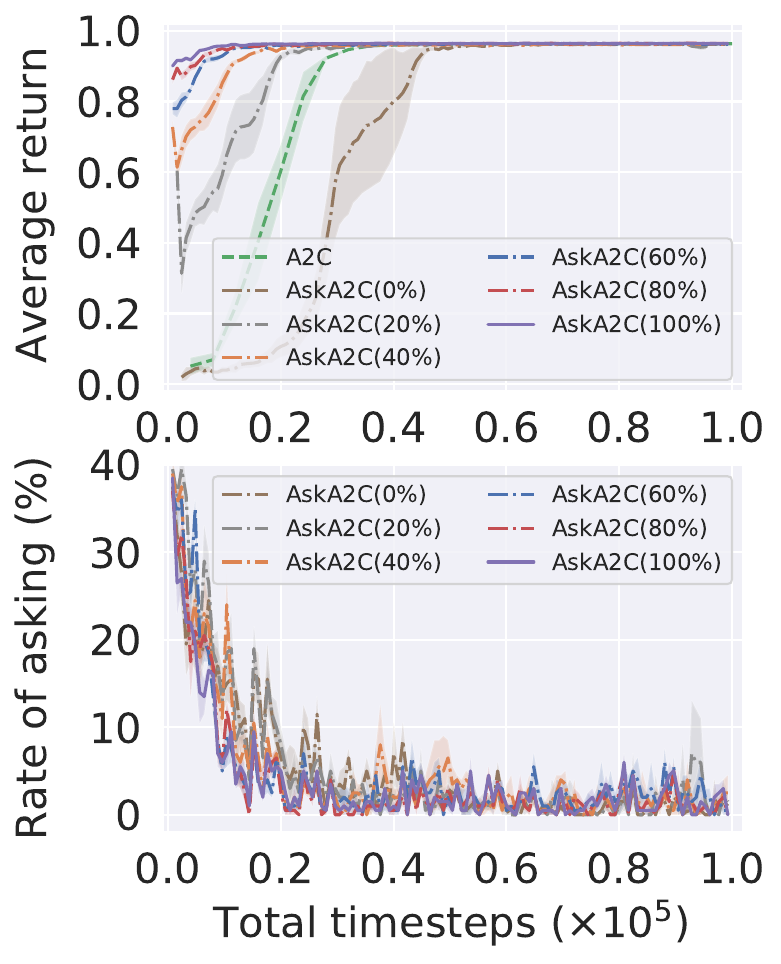}}
    \caption{Comparing the Ask-AC framework under the feedback of advisor experts with different accuracies in two stationary environments. For each AskA2C($p$) agent, we define that the interactive system has a probability $p$ of returning an advisor action and a probability $1 - p$ of returning a random action when agent seeks feedback. We set that the probability $p \in \{0\%,20\%,40\%,60\%,80\%,100\%\}$.}
  \label{fig:amateur}
\end{figure}

The interactive system simulates advisor error in the robustness analysis experiment by returning a random action with a certain probability when the agent seeks feedback. As shown in Figure~\ref{fig:amateur}, when advisor feedback is not always correct, our method demonstrates solid robustness: it yields results superior or at least on par with the baseline method when the feedback accuracy of advisor experts is greater than $40\%$. With the increase of feedback accuracy, our method can effectively use advisor feedback to improve learning efficiency. Moreover, even when the feedback accuracy is low, our method is robust enough to avoid being misled by the wrong feedback. Though the action requester does not directly check the correctness of advisor feedback, it endows the agent with an initiatively-asking action, which is learnable by the traditional reinforcement learning method. Loosely speaking, even if the advisor feedback is not always correct, the agent can benefit from it as long as the expected return of the initiatively-asking action is sufficiently high. On the other hand, if the agent finds that seeking advisor feedback cannot bring benefits, its asking ability will be gradually inhibited by the traditional reinforcement learning loss and ask loss.

\begin{figure}[!t]
  \centering
    \subfloat[The human performance of different human advisors]{\includegraphics[scale=0.32]{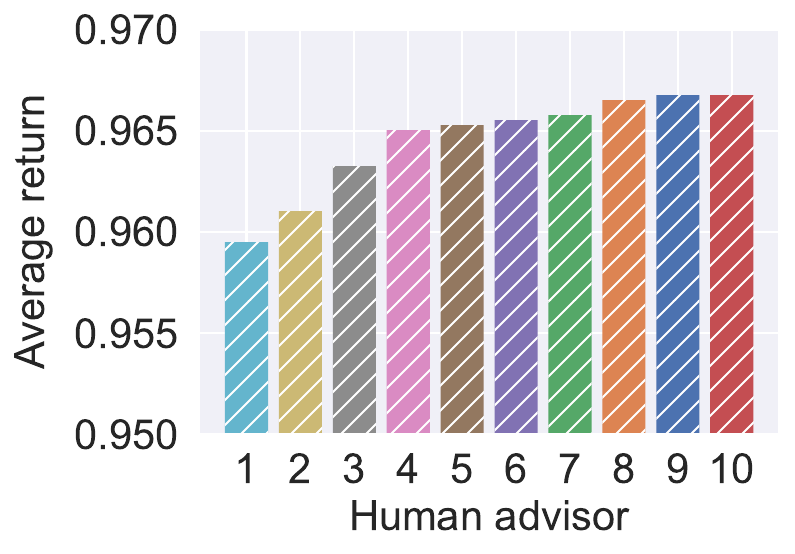}}\;
    \subfloat[The agent performance with different human advisors]{\includegraphics[scale=0.32]{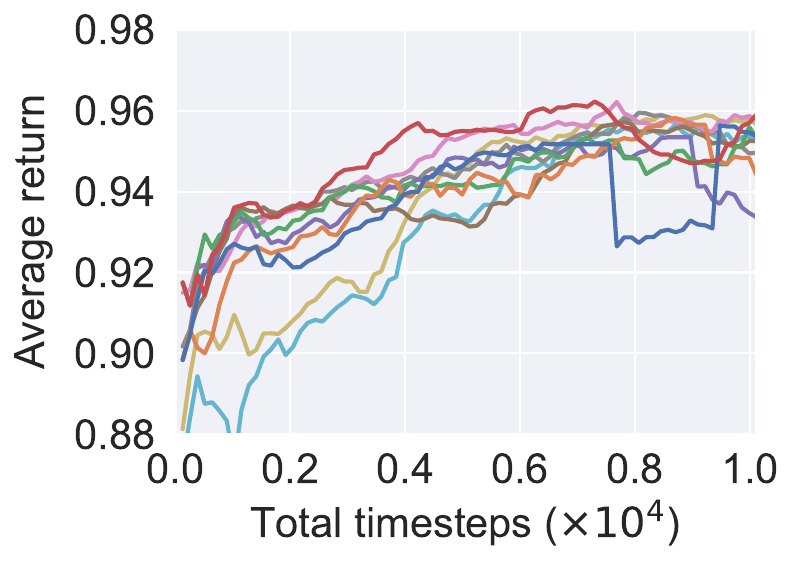}}
    \caption{Comparing the Ask-AC framework under the feedback of different human advisors in the DoorKey~($S=6\times6$) environment. The results of different human advisors are colored differently, where the corresponding colors are kept consistent between (a) and (b).}
  \label{fig:human}
\end{figure} 

\begin{figure}[!t]
  \centering
  \includegraphics[scale=0.32]{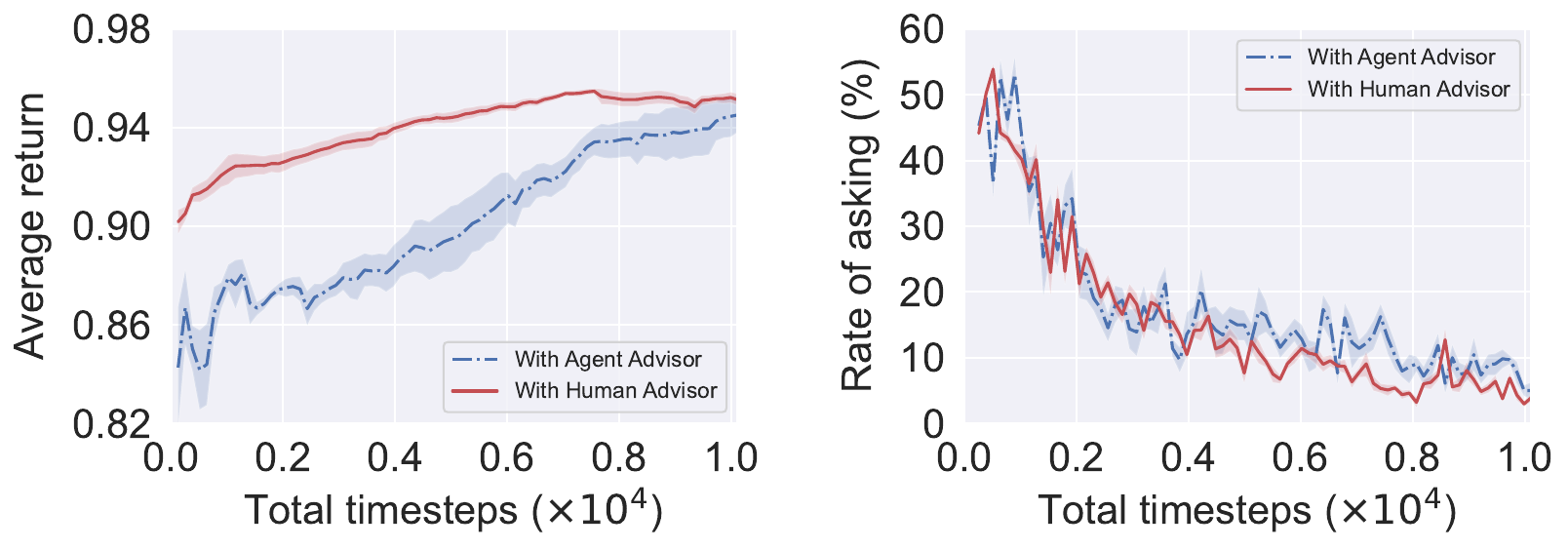}
    \caption{Comparing the Ask-AC framework under the feedback of agent advisors and human advisors in the DoorKey~($S=6\times6$) environment.}
  \label{fig:agentVShuman}
\end{figure} 

\subsection{Human Advisor Study}

To further test the practical effectiveness of our Ask-AC framework, we additionally introduce human advisors to assist agent training. A total of 10 people were recruited in our experiments, each paid ¥50 per hour. The entire experimental procedure was recorded. All participants used the same desktop computers to complete the study, where the random seeds of the environment were fixed to ensure comparability. We use the AskPPO method and adopt the DoorKey environment to conduct experiments. Our study includes two processes: (1)~Testing the performance of human advisors in playing games. (2)~Testing the performance of training agents under the feedback of human advisors.

Figure~\ref{fig:human}(a) demonstrates the game-playing performance of different human advisors.
The average test return of human advisors is $0.965$. The training agent performance under the feedback of different human advisors is shown in Figure~\ref{fig:human}(b). The results suggest that the agents assisted by human advisors with lower performance levels often performed poorly in the early stages of training. Nevertheless, all human advisors can successfully help agents to achieve the promising performance. Moreover, it is an interesting finding that human concentration often decreases as the test progresses. Thus, some human advisors may give agents several wrong feedback, resulting in a slight performance degradation of the training agents in the later stage. Figure~\ref{fig:agentVShuman} reports the experimental results of the Ask-AC framework under the feedback of agent advisors and human advisors. The average test return of agent advisors is $0.940$, which is lower than $0.965$ of human advisors. We observe that the Ask-AC framework with the human advisors outperforms the one with the agent advisors not only in the learning efficiency but also the training stability.

\section{Conclusion}

In this paper, we introduce Ask-AC, a novel interactive reinforcement learning framework that enables the learner to initiatively seek advisor assistance through a bidirectional learner-initiative mechanism. The proposed framework can be readily applied to various existing actor-critic architectures and handle discrete action-space tasks. Experimental results on both stationary and non-stationary environments show that, our framework significantly improves the learning efficiency of the agent and achieves performance comparable to those requiring constant advisor monitoring during training. Furthermore, additional analysis experiment demonstrates the robustness and effectiveness of our framework under the guidance of advisers with various operation levels.

\textbf{Limitations.} Currently, the proposed Ask-AC framework only considers the discrete control tasks that allow delayed responses, while many real-world problems involving continuous control often need real-time advisor guidance. However, implementing advisor-in-the-loop real-time control is more challenging in terms of designing advisor feedback, because Ask-AC requires advisors to give detailed actions. In complex control tasks, it is difficult for human advisors to directly provide agents with precise actions in real time, especially for continuous control. On the contrary, human advisors can easily assign a preference to a trajectory pair of agents, which implies the desired behavior. For example, in a driving task, human advisors often prefer to provide a high-level instruction~(\textit{i.e.} turn left or turn right) instead of a steering-wheel angle. Thus, our future work will extend the proposed Ask-AC framework to utilize human preference and provide a more efficient advisor-in-the-loop framework.

\ifCLASSOPTIONcaptionsoff
  \newpage
\fi

\bibliographystyle{IEEEtranN}
\bibliography{ref}

\end{document}